\documentclass{article}

\usepackage[main, final]{neurips_2026}

\usepackage[utf8]{inputenc} % allow utf-8 input
\usepackage[T1]{fontenc}    % use 8-bit T1 fonts
\usepackage{hyperref}       % hyperlinks
\usepackage{url}            % simple URL typesetting
\usepackage{booktabs}       % professional-quality tables
\usepackage{amsfonts}       % blackboard math symbols
\usepackage{nicefrac}       % compact symbols for 1/2, etc.
\usepackage{microtype}      % microtypography

\usepackage{amsmath}
\usepackage{amssymb}
\usepackage{mathtools}
\usepackage{amsthm}

\usepackage{graphicx}
\usepackage{wrapfig}
\usepackage{subcaption}
\usepackage[capitalize,noabbrev]{cleveref}
\usepackage{threeparttable}

\usepackage{algorithm}
\usepackage{algorithmic}

\usepackage{multirow}
\usepackage{microtype}
\usepackage{xcolor}         % colors
\usepackage{listings}
\usepackage[most]{tcolorbox}

\lstdefinestyle{py}{
  language=Python,
  basicstyle=\ttfamily\small,
  keywordstyle=\color{blue!70!black}\bfseries,
  stringstyle=\color{green!40!black},
  commentstyle=\color{green!40!black}\itshape,
  showstringspaces=false,
  breaklines=true,
  breakatwhitespace=true,
  tabsize=2,
  numbers=none
}

\newtcblisting{pycode}{
  listing only,
  listing options={style=py},
  colback=white,
  colframe=black!60,
  boxrule=0.6pt,
  arc=0pt,
  left=10pt,
  right=10pt,
  top=8pt,
  bottom=8pt,
  breakable
}

\title{Learning to Solve Compositional Geometry\\ Routing Problems}

\author{
  Mingfeng Fan \\
  National University of Singapore\\
  \texttt{ming.fan@nus.edu.sg} \\
  \And
  Jianan Zhou\thanks{Corresponding author.}\\
  Nanyang Technological University \\
  \texttt{jianan004@e.ntu.edu.sg} \\
  \And
  Jiaqi Cheng$^*$ \\
  Central South University \\
  \texttt{chengjq0705@gmail.com} \\
  \And
  Yifeng Zhang \\
  National University of Singapore \\
  \texttt{yifeng@u.nus.edu} \\
  \And
  Jie Zhang \\
  Nanyang Technological University\\
  \texttt{zhangj@ntu.edu.sg} \\
  \And
  Guillaume Adrien Sartoretti \\
  National University of Singapore\\
  \texttt{guillaume.sartoretti@nus.edu.sg} \\
}

\begin{document}

\maketitle

\begin{abstract}
We study the Compositional Geometry Routing Problem (CGRP), a unified superclass of traditional routing problems that covers point-only, line-only, area-only, and arbitrary hybrid task geometries, providing a broad abstraction for real-world routing scenarios. Beyond standard point-based routing, CGRP with non-point tasks can be inherently asymmetric, tightly coupled travel routes with the intrinsic path, and enlarges the action space with numerous feasible yet often irrelevant options, thereby posing significant challenges for both representation learning and decision-making. To address these challenges, we propose DiCon, a differential attention–assisted solver with contrastive learning, as a plug-and-play framework that tackles the problem from two complementary angles. First, we introduce a differential attention mechanism that actively suppresses the probability mass on less competitive candidate actions. Second, we design a double-level contrastive learning objective to promote robust global instance representations and regularize geometry-aware task representations. Extensive experiments demonstrate that DiCon achieves strong performance, broad versatility, and superior generalization across diverse CGRP instances with different compositions.
\end{abstract}

\section{Introduction}
\label{sec:intro}

% Routing problems, such as the Traveling Salesman Problem (TSP), have been widely studied due to their significance in practical applications such as logistics~\cite{veenstra2017pickup}, robotic inspection~\cite{asghar2024multi}, and disaster response~\cite{wang2016novel}. A common modeling choice represents targets as nodes and optimizes a tour or a set of routes that visits them~\cite{flood1956traveling}. However, this abstraction can be misaligned with real-world missions, where critical objectives often take the form of extended structures (e.g., cracks along seams, power lines, levees)~\cite{nekovavr2021multi} or spatial regions that must be searched or covered (e.g., collapsed buildings, farmland)~\cite{plessen2025path}. In such settings, the decision extends beyond which target to visit next to include \emph{how to enter and exit each target} in a manner that interacts with the global route.
Routing problems have long served as fundamental models for spatial decision-making, with broad applications in logistics~\cite{veenstra2017pickup}, robotic inspection~\cite{asghar2024multi}, and disaster response~\cite{wang2016novel}. Classical formulations usually assume a specific service geometry. For example, node-routing problems, such as the Traveling Salesman Problem (TSP), represent each task as a discrete point and optimize the visiting order among these points~\cite{flood1956traveling}. Line- or arc-routing problems focus on extended structures, such as roads, cracks, and power lines~\cite{nekovavr2021multi,agarwal2024line}. Area-routing or coverage-routing problems further require robots to service spatial regions, such as fields, buildings, or search zones~\cite{plessen2025path, zhang2025multi}.
However, many real-world missions cannot be naturally described by a single type of service geometry. Robotic inspection may require covering spatial regions while tracking crack-like linear defects~\cite{veeraraghavan2024complete}.
Similarly, search-and-rescue, infrastructure maintenance, and environmental surveying may involve point tasks, line tasks, area tasks, or arbitrary combinations of them~\cite{gao2021multi,zhu2018multi}. Treating these cases as separate routing problems limits the flexibility of solver design and often requires geometry-specific adaptation. 

\begin{figure}
    \vspace{-1mm}
    \centering
    \includegraphics[width=0.9\columnwidth]{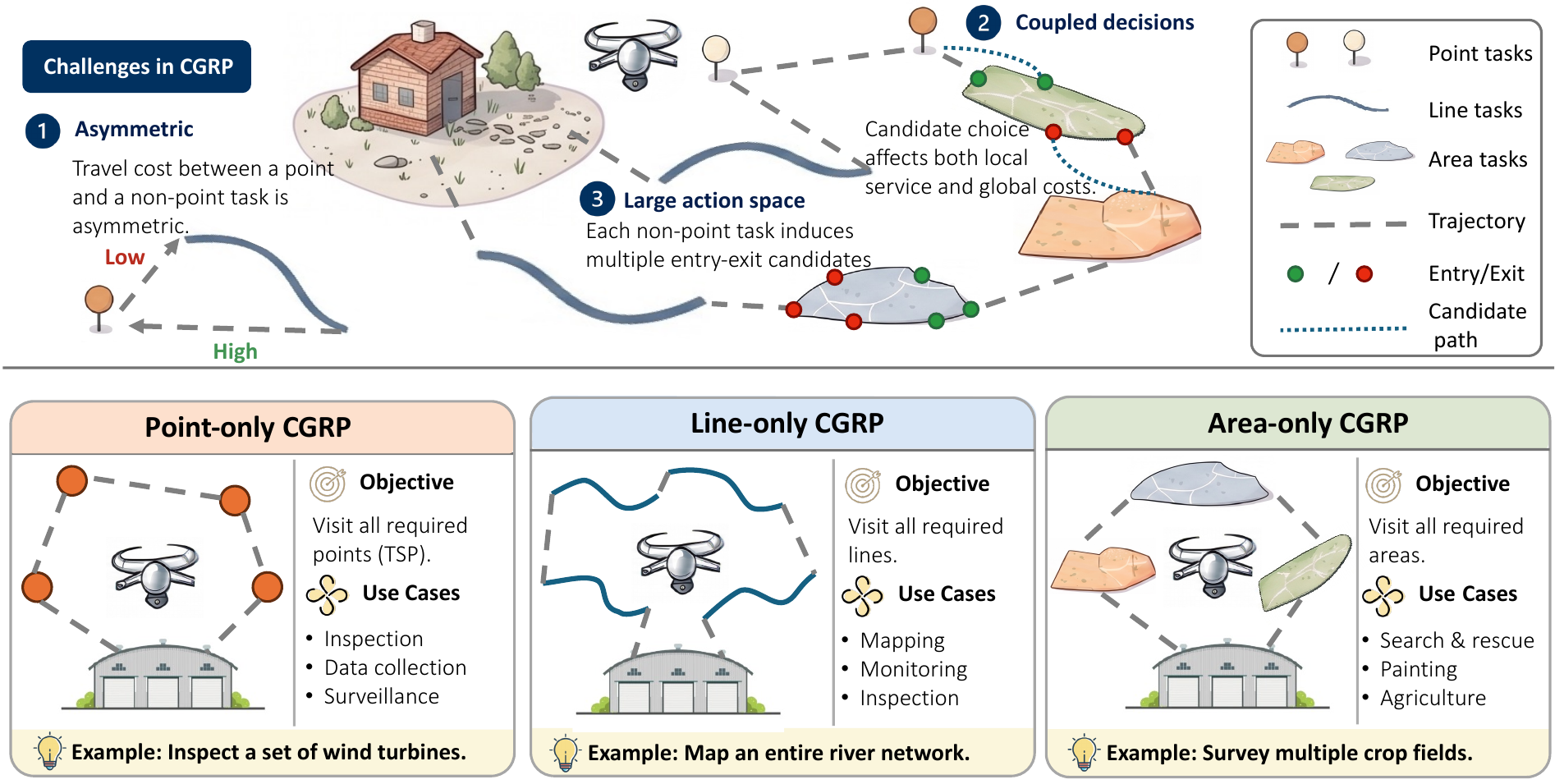}
    \vspace{-1mm}
    \caption{The characteristics of CGRP.}
    \label{fig:challenge}
    \vspace{-6mm}
\end{figure}

To provide a common abstraction for homogeneous and mixed-geometry routing, we study the Compositional Geometry Routing Problem (CGRP), a unified formulation over point, line, and area tasks in a two-dimensional workspace. CGRP subsumes several classical routing settings while allowing arbitrary mixtures of service geometries within a single instance. Unlike point tasks, non-point tasks impose intrinsic service requirements, such as traversing a line segment or covering a region with a feasible pattern, and often admit multiple valid entry--exit pairs. The unified nature of CGRP introduces several challenges beyond classical node-centric routing, as illustrated in Fig.~\ref{fig:challenge}. First, the problem with non-point tasks can become asymmetric, since the cost of traveling from a preceding task to a non-point task generally differs from the cost of traveling from that non-point task back to the preceding task. Second, route planning and intrinsic service paths are tightly coupled: selecting an entry--exit pair for a line or area task is both a local servicing decision and a global routing decision, because it changes the connection lengths to preceding and succeeding tasks. Third, the action space expands substantially, as each line or area task gives rise to multiple feasible candidates, many of which may be geometrically valid but strategically irrelevant. 

% To capture these practical requirements, we study the Compositional Geometry Routing Problem (CGRP), which generalizes node-only routing by allowing heterogeneous spatial tasks in a two-dimensional workspace, including point, line, and area tasks~\cite{zhu2018multi}. 
% Each non-point task induces an intrinsic service path (e.g., traversing a line segment or covering an area according to a feasible pattern), and typically admits multiple feasible entry–exit pairs. These characteristics make CGRP fundamentally different from classical node-centric routing in several key aspects, as illustrated in Fig. \ref{fig:challenge}.
% First, the problem becomes asymmetric, as the cost of traveling from a preceding task to a non-point task generally differs from the cost of traveling from that non-point task back to the preceding task.
% Second, route planning and the intrinsic paths of non-point tasks are tightly coupled, as local servicing decisions (i.e., selecting entry–exit pairs) directly affect global travel costs by altering the connection lengths to preceding and succeeding tasks.
% Third, the action space expands substantially, since each line or area task gives rise to numerous feasible (yet often irrelevant) candidates, which amplifies the combinatorial complexity.

While traditional exact and heuristic methods exist, they often require substantial domain knowledge, and efficiently finding near-optimal solutions remains challenging, particularly for large-scale instances~\cite{lin1973effective, karapetyan2011lin,helsgaun2000effective,dorigo2002ant}. Recent neural combinatorial optimization (NCO) approaches have shown strong empirical performance on a variety of routing problems~\cite{berto2025rl4co, wu2024neural}. However, most neural solvers are tailored to node-centric formulations with relatively structured action spaces. When applied to CGRP, they encounter two recurring challenges: 
1) \emph{Representation:} the model must encode task geometries and their service implications in a manner that generalizes across single-geometry instances, hybrid compositions, and diverse spatial layouts.
2) \emph{Decision-making:} the model needs to assign probability mass over a substantially larger and noisier candidate set, where many options are geometrically feasible but strategically less competitive. These challenges are particularly pronounced under compositional generalization, where test instances present novel combinations of task types, scales, and geometric configurations that were not jointly observed during training.

To address these challenges, we introduce \underline{Di}fferential attention-assisted solver with \underline{Con}trastive learning (DiCon), a plug-and-play framework designed to augment neural routing backbones with enhanced representation learning and decision-making for compositional geometry tasks. DiCon comprises two complementary mechanisms. 
First, a differential attention (DA) mechanism is introduced to suppress probability mass assigned to less competitive candidate actions, thereby mitigating over-exploration of geometrically feasible yet strategically weak entry–exit choices.
Second, we design a double-level contrastive learning (CL) objective that operates at two granularities: an instance level, which promotes robust global representations across diverse task compositions, and an intra-task level, which reinforces semantic coherence within each task while preserving representational diversity across different tasks.
Our contributions are summarized as follows:
\begin{itemize}
    \item  We study CGRP, which more faithfully captures the complexity of diverse real-world applications. We formulate CGRP solving as a Markov Decision Process (MDP) and propose a plug-and-play framework that enables neural solvers to tackle CGRP effectively.
    
    \item We identify key challenges in neural routing backbones and propose a differential attention mechanism together with a double-level CL objective to enhance both representation learning and decision-making for compositional geometry tasks.
    
    \item We conduct experiments on diverse compositional geometry settings and integrate DiCon into different neural routing backbones. Empirical results show that DiCon achieves strong performance, broad versatility, and robust compositional generalization.
\end{itemize}

\section{Related Work}

\subsection{Neural Routing Solver}
Pointer Networks~\cite{vinyals2015pointer} pioneered neural heuristics for TSP by framing solution construction as a sequence-to-sequence prediction task with attention. While effective, this approach relies on costly supervised labels. \cite{bello2016neural} replaced supervision with reinforcement learning (RL), enabling direct optimization of tour length. The Attention Model (AM)~\cite{kool2018attention} further advanced this line of work through a Transformer-style encoder–decoder that constructs solutions autoregressively. Exploiting the head–tail symmetry of solutions, POMO~\cite{kwon2020pomo} introduced parallel rollouts with strong data augmentation, substantially outperforming AM and inspiring many subsequent studies~\cite{kim2022sym,HottungKT22,grinsztajn2023winner,gao2023towards, huang2025rethinking}. 
On the other hand, several studies propose an alternative architectural paradigm, the Light Encoder-Heavy Decoder (LEHD)~\cite{drakulic2023bq,luo2023neural}, which is typically trained with supervised labels. To reduce the dependence on expensive labels, subsequent works~\cite{pirnay2024selfimprovement,luo2025boosting} train LEHD models without labels through self-improving learning.
Beyond these advances, more recent efforts have increasingly focused on improving generalization to large-scale instances~\cite{li2021learning,zong2022rbg,hou2023generalize,ye2024glop,zheng2024udc}, robustness under distribution shifts~\cite{bi2022learning}, and broader problem variants~\cite{zhou2024mvmoe,berto2025routefinder,drakulic2025goal}.

In this paper, we focus on the above construction-based solvers and refer interested readers to a comprehensive survey~\citep{wu2024neural}. Despite recent extensions, most neural routing solvers remain tailored to node-centric formulations, including asymmetric variants with direction-dependent pairwise costs~\cite{kwon2021matrix,li2025heterogeneous}, arc routing problems that require servicing edges rather than nodes~\cite{jia2025neural}, and generalized TSP settings in which each cluster provides multiple candidate nodes~\cite{pmlr-v305-cheng25a}. In contrast, CGRP covers compositional geometry tasks, including point-only, line-only, area-only, and hybrid geometries. By incorporating non-point tasks with intrinsic service paths and multiple entry--exit candidates, CGRP defines a more challenging, general, and practically relevant problem setting.

\subsection{Contrastive Learning for NCO}
Contrastive learning (CL) is a self-supervised learning paradigm that acquires transferable representations by encouraging agreement between positive samples while pushing apart representations of negative samples, thereby exploiting structural similarity in unlabeled data. It has been extensively studied in computer vision~\cite{he2020momentum,tian2020contrastive,chuang2020debiased} and natural language processing~\cite{mikolov2013distributed,giorgi2021declutr}, where it consistently improves representation quality and downstream generalization. 
A representative method is SimSiam~\cite{chen2021exploring}, which adopts a simple Siamese architecture with asymmetric prediction and a stop-gradient operation to prevent representation collapse, avoiding the need for explicit negative pairs or large memory banks.

In NCO, CL has emerged as an effective self-supervised paradigm for enhancing representation learning and robustness, particularly under limited supervision or distribution shifts. \cite{duan2022augment} provided an early study, demonstrating that CL benefits combinatorial problems when augmentations preserve feasibility and cost structure, while naive perturbations can be detrimental. Building on this insight, \cite{jiang2023multi} proposed a multi-view graph contrastive framework for vehicle routing, contrasting complementary instance views to promote invariant representations. Beyond routing, \cite{huang2024contrastive} introduced a contrastive predict-and-search framework for mixed-integer linear programs, aligning predictions with search trajectories to guide solution discovery. CL has also been integrated with RL: \cite{yuan2022rl} combined contrastive objectives with policy learning to strengthen state representations and accelerate training. More recently, \cite{yuanoptfm} scaled CL to a hierarchical pretraining regime using a multi-view graph Transformer, enabling transfer across diverse combinatorial optimization tasks. In addition, \cite{guoconrep4co} aligned multiple graph decision problems with their translated Boolean satisfiability forms through a cross-modal contrastive pretraining paradigm.
In this paper, we design a double-level CL objective to enhance the representation learning for compositional geometry tasks.

\section{Problem Statement}

\textbf{CGRP.} The CGRP generalizes classical routing problems by covering point-only, line-only, area-only, and arbitrary hybrid task geometries within a unified formulation. Non-point tasks involve intrinsic service paths and multiple candidate entry-exit configurations. A \emph{point task} is considered served when the agent passes through its center. A \emph{line task} must be physically traversed along its segment, offering two candidate entry-exit pairs corresponding to traversal in either direction. For an \emph{area task}, the agent must execute a coverage path under detection range $\gamma$, without assuming a specific coverage pattern. For each area task, we retain the $\omega$ shortest feasible coverage paths as candidate service options, each associated with a corresponding entry--exit pair.

% \colorr{For an \emph{area task}, the agent must perform a zigzag coverage path with detection range $\gamma$, aligned with the longer side to minimize traversal cost. Each area task therefore provides four candidate entry-exit pairs located at the endpoints of its shorter sides.}

Formally, let $n_a$, $n_l$, and $n_p$ denote the number of area, line, and point tasks, respectively, where each of them can be zero to allow point-only, line-only, area-only, and hybrid instances. We define the full candidate set as \( \mathcal{V} = \{0\} \cup \mathcal{V}_a \cup \mathcal{V}_l \cup \mathcal{V}_p \), where $0$ is the depot. The area candidate set is \( \mathcal{V}_a = \{1, \ldots, \omega\cdot n_a\} \), with $\omega$ candidates per area task. The line candidate set is \( \mathcal{V}_l = \{\omega\cdot n_a + 1, \ldots, \omega\cdot n_a + 2n_l\} \), and the point task set is \( \mathcal{V}_p = \{\omega\cdot n_a + 2n_l + 1, \ldots, \omega\cdot n_a + 2n_l + n_p\} \). 
In this paper, we treat candidate nodes within the same task (e.g., the $\omega$ entry–exit pairs associated with a single area task) as \emph{semantically equivalent}.
Each candidate is associated with an entry--exit coordinate pair. For each line or area candidate $i \in \mathcal{V}_a \cup \mathcal{V}_l$, we represent this pair as $L_i=\{L_i^+, L_i^-\}$, where \( L_i^+ \) and \( L_i^- \) denote the entry and exit locations, respectively. Additionally, each such candidate $i \in \mathcal{V}_a \cup \mathcal{V}_l$ is assigned a service cost \( c_i \), corresponding to the segment length for line tasks or the coverage path length induced by the geometry of the associated area task. For other candidates $ i\in \{0\} \cup \mathcal{V}_p $, the entry and exit locations coincide and the service cost $c_i=0$.

The agent must construct a tour that starts and ends at the depot, selects exactly one entry-exit pair for each non-point task, and visits all tasks exactly once.
Let \( \pi = (\pi_0, \pi_1, \dots, \pi_N) \), where $N=n_a+n_l+n_p$, $\pi_i \in \mathcal{V}$, and \( \pi_0 = 0 \), denote a sequence of selected candidates (with exactly one candidate chosen per task). 
Let \( d_{\pi_i\pi_j} \) denote the travel cost from candidate \( \pi_i \) to candidate \( \pi_j \). 
Importantly, \( d_{\pi_i\pi_j} \ne d_{\pi_j\pi_i} \) in general due to the asymmetric nature of line and area tasks. For example, if \( \pi_i \in \mathcal{V}_p \) and \( \pi_j \in \mathcal{V}_l \), then $d_{\pi_i\pi_j} = \| L_i - L_j^+ \|_2 $ and \( d_{\pi_j\pi_i} = \| L_j^- - L_i \|_2 \). 

The total cost consists of the sum of all inter-task travel distances, including the return to the depot, as well as the intrinsic service costs associated with each line and area task. The objective is given by:
\begin{equation}
    \min_{\pi} f(\pi)=\sum_{i=0}^{N-1} d_{\pi_i\pi_{i+1}} + d_{\pi_{N}\pi_0} + \sum_{j=0}^{N} c_{\pi_j}.
\end{equation}

This formulation captures the key characteristics of CGRP, including task-specific geometries, asymmetric routing, intrinsic service paths, and combinatorial candidate selection, while retaining classical routing problems as special cases.
% rendering it substantially more challenging yet practically relevant than classical routing problems, which can be viewed as special cases of CGRP.

\textbf{MDP Formulation.} To solve CGRP with RL-based neural policies, we formulate it as an MDP \( \mathcal{M} = (\mathcal{S}, \mathcal{A}, \mathcal{P}, \mathcal{R}) \), where $\mathcal{S}$ is the state space, $\mathcal{A}$ is the action space, $\mathcal{P}$ denotes the transition dynamics, and $\mathcal{R}$ is the reward function.

\begin{itemize}
    \item \textbf{State.} A state \( s_t \in \mathcal{S} \) encodes the current partial solution together with the input instance \( g\in \mathcal{G} \). The initial state \( s_0 \) contains only the depot, indicating the agent's starting position. The terminal state \( s_T \) corresponds to a complete solution \( \pi = (\pi_0, \pi_1, \dots, \pi_T) \), where \( T = N \), and all tasks have been visited exactly once.
    
    \item \textbf{Action.} The action space \( \mathcal{A} =\mathcal{V}\) consists of the depot \( 0 \), all point tasks \( \mathcal{V}_p \), and all candidate entry-exit pairs of line and area tasks in $\mathcal{V}_l$ and $\mathcal{V}_a$.
    An action \( a_t \in \mathcal{A} \) selects the next task to visit, along with its entry-exit pair if applicable. 
    
    \item \textbf{Transition.} The next state $s_{t+1}$ is derived deterministically from $s_t$ by the chosen action $a_t$.
    Once a candidate of a non-point task is selected as $a_t$, all other candidates corresponding to that task are marked as visited. 
    The depot remains masked until all tasks are visited.
    
    \item \textbf{Reward.} The agent receives zero reward at each intermediate step $t<T$ and a terminal reward at $s_T$ equal to the negative total cost of the constructed tour $\mathcal{R}(s_T, \cdot) = -f(\pi)$.
    
    \item \textbf{Policy.} The policy \( p_\theta \), parameterized by \( \theta \), constructs a complete solution autoregressively, and its probability can be factorized via the chain rule as:
    \begin{equation}
         p_\theta(\pi | g) = \prod_{t=0}^{T} p_\theta(\pi_t | \pi_{<t}, g).
    \end{equation}
    With a slight abuse of notation, we use \( \pi_t \) to denote the action selected at step \( t \), and \(\pi_{<t} \) to represent the partial solution up to step \( t-1 \).
\end{itemize}

\section{Methodology}
\begin{figure*}
    \vspace{1mm}
    \centering
    \includegraphics[width=\linewidth]{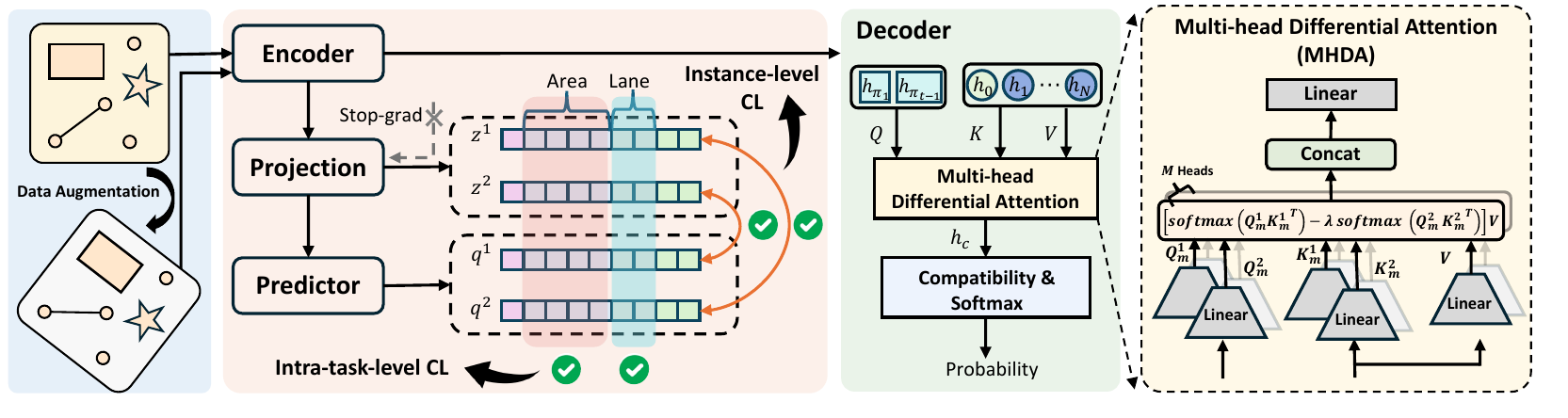}
    \caption{An overview of DiCon. DiCon incorporates two mechanisms: 1) a multi-head differential attention module that replaces the standard MHA in the decoder to facilitate decision-making, and 2) a double-level contrastive objective to enhance representation learning.}
    \label{fig:method}
    \vspace{-2mm}
\end{figure*}

\subsection{Overview} \label{sec: method_overview}

%Most neural routing solvers adopt a Transformer-style architecture consisting of a multi-layer encoder and a single-layer decoder. To extend these architectures for solving CGRP, we introduce a unified encoding scheme that handles all three task geometries. Given an instance \( g \in \mathcal{G} \), we extract three sets of raw features: the candidate location \( L_i \) for each \( i \in \mathcal{V} \), a geometric center \( G_i \in \mathbb{R}^2 \) and a task-type indicator \( I_i \) for each non-depot candidate \( i \in \mathcal{V} \setminus \{0\} \) where \( I_i \in \{0, 1, 2\} \) denotes area, line, and point tasks respectively. We first build the complete locations features $L_i^\mathcal{T} = \{L_i, L_i\} \in \mathbb{R}^4$ if $i \in \mathcal{V}_p$ else $L_i^\mathcal{T} = L_i = \{L_i^+, L_i^-\}$. Then, we construct the initial task embeddings $H_\mathcal{T}^0=\{h_1^0, \cdots, h_N^0\}$ for each non-depot candidate \( i \in \mathcal{V} \setminus \{0\} \) by summing of the linear projection of complete locations $L_i^\mathcal{T}$, geometric center $G_i$, and task-type indicator \( I_i \). Similar, we construct the initial depot embedding $H_d^0$ by linear transformation of the depot location $L_0$. Finallyl, we obtain the initial embeddings by concatenation initial task embeddings and initial node embeddings such as $H^0=Concat(H_d^0, H_\mathcal{T}^0) \in \mathbb{R}^{(N+1)\times d}$ where $d$ is the hidden dimension and $Concat$ is the concatenation operation.

Most neural routing solvers adopt a Transformer-style architecture consisting of a multi-layer encoder and a single-layer decoder. To extend such architectures to the CGRP, we propose a unified encoding scheme capable of handling different task geometries. Its effectiveness is validated in Appendix~\ref{sec:unified_encoding}. Given an instance \( g \in \mathcal{G} \), we extract three sets of raw features: 
1) the coordinate pair \( L_i \in \mathbb{R}^4 \) for each \( i \in \mathcal{V} \);
2) a geometric anchor \( G_i \in \mathbb{R}^2 \) for each non-depot candidate \( i \in \mathcal{V} \setminus \{0\} \), representing either the task location (for point tasks) or the geometric center (for line and area tasks); and 
3) a task-type indicator \( I_i \in \{0, 1, 2\} \) representing area, line, and point tasks, respectively, also defined for each \( i \in \mathcal{V} \setminus \{0\} \).
% We construct a complete location representation \( L_i^\mathcal{T} \in \mathbb{R}^4 \) as follows: for point tasks \( i \in \mathcal{V}_p \), we duplicate the location \( L_i^\mathcal{T} = [L_i, L_i] \); for non-point tasks \( i \in \mathcal{V}_l \cup \mathcal{V}_a \), we use their entry-exit pair \( L_i^\mathcal{T} = [L_i^+, L_i^-] \). 
The initial embedding for each non-depot node is then given by:
\begin{equation}
    h_i^0 = W_L L_i + W_G G_i + W_I I_i + b,
\end{equation}
where \( W_L, W_G, W_I \) are learnable projection matrices and \( b \) is a shared bias term. The initial embedding of the depot is computed separately as $h_0^0 = W_d L_0 + b_d$, where $W_d$ and $b_d$ are learnable parameters.
Finally, we obtain the full input sequence as: $H^0 = [h_0^0, h_1^0, \ldots, h_{|\mathcal{V}|}^0] \in \mathbb{R}^{(|\mathcal{V}|+1) \times d}$,
where \( d \) is the hidden dimension and \( [\cdot] \) denotes the row-wise concatenation.
This unified encoding allows the encoder to reason over compositional geometry tasks within a shared representation space while preserving the semantic and geometric diversity essential for CGRP. The $\mathbb{L}$-layer encoder transforms the initial embeddings $H^0$ into task-aware representations $H^\mathbb{L}$. Conditioned on the output of the encoder and the evolving state $s_t$, the decoder is repeatedly invoked to select actions in an autoregressive manner. 

As illustrated in Fig.~\ref{fig:method}, DiCon introduces two dedicated components to address the intrinsic challenges of CGRP, namely, the abundance of feasible yet less competitive candidate actions and the compositional geometric structure of tasks.
First, we incorporate a DA mechanism into the decoder to suppress probability mass assigned to strategically weak candidates, thereby promoting more focused and effective decision-making. Second, to improve representational capacity for compositional geometry reasoning, we introduce a double-level CL objective: an instance-level CL loss that promotes robust global representations across different task compositions, and an intra-task CL loss that reinforces semantic coherence within the same task while preserving inter-task diversity. Notably, DiCon is a generic, plug-and-play framework that can be seamlessly integrated into existing neural routing solvers without architectural redesign. 

\subsection{Differential Attention Mechanism}
Most neural routing solvers employ a single-layer decoder composed of a multi-head attention (MHA) sublayer followed by a compatibility sublayer. Given the encoder output $H^\mathbb{L}=\{h_0^{\mathbb{L}},\ldots,h_{|\mathcal{V}|}^{\mathbb{L}}\}$ and contextual features at step $t$, such as the embedding of the first selected node $h_{\pi_1}$ and the most recently visited node $h_{\pi_{t-1}}$, the decoder first constructs query, key, and value representations to compute a context embedding $h_c$ via MHA.
This context embedding is then passed through a compatibility sublayer to produce a probability distribution over all available actions. The details of the model forward pass are provided in Appendix~\ref{appendix_model}.
To enhance robustness in the presence of numerous feasible yet less competitive actions, a common challenge in CGRP, we integrate a DA mechanism into the MHA sublayer of the decoder. This mechanism performs attention denoising by suppressing the influence of low-saliency candidate actions, enabling the decoder to focus more effectively on critical decision points \cite{ye2025differentialtransformer}. 
% Importantly, DA is incorporated without modifying the backbone architecture of the neural solver, thereby maintaining efficiency and compatibility. 
Through this targeted integration, DiCon improves the decoder’s discrimination and decision-making capability in compositional geometry routing settings.
% DiCon improves the discrimination and selection capability of the decoder in compositional geometry routing contexts.

Specifically, we partition the query and key vectors into two groups and compute two separate softmax attention maps for each head in the MHA sublayer. Formally, for head $m \in [1, \cdots, M]$ with total heads number $M$:
\begin{equation}
    Q_m^1 = [h_{\pi_1}, h_{\pi_{t-1}}]W_{q_m}^1,\ Q_m^2 = [h_{\pi_1},h_{\pi_{t-1}}]W_{q_m}^2,
\end{equation}
\begin{equation}
    K_m^1 = H^\mathbb{L}W_{k_m}^1,\ K_m^2 = H^\mathbb{L}W_{k_m}^2,
\end{equation}
where $W_{q_m}^1, W_{q_m}^2, W_{k_m}^1, W_{k_m}^2$ are learnable parameters. Then we obtain the softmax logits of the two groups of query and key as follows:
\begin{equation}
    A_m^1 = \text{softmax}(\frac{Q_m^1{K_m^1}^T}{\sqrt{d/m}}),\ A_m^2 = \text{softmax}(\frac{Q_m^2{K_m^2}^T}{\sqrt{d/m}}).
\end{equation}
Then the result of subtracting these two maps is regarded as attention scores: $s_m = (A_m^1 - \lambda_m A_m^2)V$, where $\lambda_m$ is a head-dependent learnable scalar initialized as $\lambda_0$ (see Appendix~\ref{sec:implementation}) and $V=H^\mathbb{L}W_{v_m}$ with the learnable parameter $W_{v_m}$. The DA mechanism mitigates attention noise, encouraging the model to focus on critical information (see Appendix~\ref{sec: entropy}). The outputs of all attention heads are projected to the final context embedding: $h_c = [s_1, \cdots, s_M]W_O$,
where $W_O \in \mathbb{R}^{d\times d}$ is a learnable projection matrix. 
The context embedding \(h_c\) is used to compute the compatibility scores 
\(\alpha_i = C \cdot \tanh\left( \tfrac{h_c^\top W^K h^{\mathbb{L}}_i}{\sqrt{d}} \right)\), 
where \(\alpha_i = -\infty\) if the candidate is masked, \(C\) is a clipping constant, and \(W^K\) is a learnable projection. 
The final selection probability is given by 
\(p_\theta(\pi_t \mid \pi_{<t}, g) = \operatorname{Softmax}(\alpha)\). 
During training, parallel solutions are sampled from $p_\theta(\cdot \mid g)$, while inference selects the highest-probability candidate greedily.
% and $Concat(\cdot)$ concatenates the heads together along the channel dimension. 

\subsection{Double-Level Contrastive Learning}
Most neural solvers rely on RL to learn a policy \( p_\theta \) that produces near-optimal solutions. However, the compositional geometry nature of CGRP poses representation challenges. To address this, we augment the standard RL objective with two CL objectives: an \emph{instance level} CL objective and an \emph{intra-task} CL objective. Together, these objectives enhance the representational capacity of the backbone model and enable effective reasoning over compositional geometric tasks.
%The instance-level objective promotes globally consistent representations across CGRP instances, thereby improving robustness to instance variation. In contrast, the intra-task-level objective refines the granularity of representations by distinguishing semantically distinct candidates that correspond to the same underlying task, sharpening task-specific features and facilitating more discriminative decision-making.

\noindent\textbf{Instance Level CL Loss.}
To improve the robustness of global representations for CGRP instances, we adopt a data augmentation strategy that exploits the rotation invariance of routing problems, under which the optimal solution remains unchanged by rigid transformations of the coordinate system. Specifically, given an instance \( g \in \mathcal{G} \), we generate \( \mathbb{A} \) augmented views by applying random spatial rotations and axis permutations to all geometric entities, including candidate locations \( L_i \) and geometric anchors \( G_i \). 

% \colorr{Following~\cite{kim2022sym},} 
In specific, we first sample a transformation index \( \alpha \sim \mathcal{U}(0,1) \), then apply a continuous rotation by
\begin{equation}
\phi = 
\begin{cases}
4\pi \alpha, & \alpha < 0.5, \\
4\pi (\alpha - 0.5), & \alpha \ge 0.5,
\end{cases}
\end{equation}
around the center of a unit square. Given a coordinate \( (x, y) \), the rotated coordinates are computed as:
\begin{equation}
\begin{aligned}
x' &= \cos(\phi)(x - 1/2) - \sin(\phi)(y - 1/2) + 1/2, \\
y' &= \sin(\phi)(x - 1/2) + \cos(\phi)(y - 1/2) + 1/2.
\end{aligned}
\end{equation}
Additionally, for half of the transformations (\( \alpha \ge 0.5 \)), the coordinate axes are swapped, i.e., \( (x', y') \leftarrow (y', x') \), introducing reflectional symmetry augmentation. This transformation is applied consistently to all geometric components (i.e.,  candidate locations \( L_i \) and geometric anchors \( G_i \)). Repeating this process \( \mathbb{A}-1 \) times yields a set of augmented instances \( \{g^1, \dots, g^{\mathbb{A}}\} \), where $g^1$ is the original instance.
As shown in Fig. \ref{fig:method}, the encoder processes each augmented instance to produce all candidate embeddings \( H^{\mathbb{L}}(g^\xi), \xi\in \{1, \cdots, \mathbb{A}\}\), which are mapped through a projection head \( \varphi(\cdot) \) and a predictor head \( \psi(\cdot) \) to obtain latent representations \( z^\xi = \varphi(H^{\mathbb{L}}(g^\xi)) \) and predictions \( q^\xi = \psi(z^\xi) \). Following a SimSiam-style objective~\cite{chen2021exploring}, we encourage agreement between different augmented views by maximizing the cosine similarity $\mathcal{D}$ between the predictor output and the stop-gradient projection output:
\begin{equation}
\mathcal{D}(q^\xi, z^\tau) = \frac{q^\xi}{\|q^\xi\|_2} \cdot \frac{z^\tau}{\|z^\tau\|_2},\ \xi \neq \tau.
\end{equation}
The instance-level contrastive loss is defined as:
\begin{equation} \label{eq: IL_CL}
\mathcal{L}_{\text{CL}}^{\text{ins}} = -\frac{1}{\mathbb{A}(\mathbb{A}-1)} \sum_{\xi=1}^{\mathbb{A}} \sum_{\tau \neq \xi} \mathcal{D}(q^\xi, z^\tau).
\end{equation}
%The encoder, projection head, and predictor parameters are optimized jointly with the RL objective.
\noindent\textbf{Intra-Task CL Loss.} To enable the model to distinguish between different tasks in the latent space, task representations should be well separated. However, for non-point tasks, multiple geographically distinct entry-exit pairs correspond to the same underlying task. Treating these candidates independently, without accounting for their shared semantic identity, may hinder representation learning. To address this, we introduce an intra-task contrastive objective that encourages the embeddings of entry-exit candidates belonging to the same task to be close in the latent space. Unlike instance-level contrast, which promotes global consistency, this objective refines the task-specific feature subspace by clustering candidate embeddings of the same task while preserving separability across tasks.

Specifically, given a set of augmented views \( \{g^1,\cdots,g^{\mathbb{A}}\} \) and their latent embeddings \( z^\xi \) and predictions \(q^\xi\), for \(\xi\in\{1,\cdots,\mathbb{A}\}\), we compute contrastive losses separately for area and line tasks. 
For each area task, we randomly select the prediction \(q_i^\xi\) of one candidate $i$ from view \(\xi\) as the query and treat the remaining $\omega-1$ semantically equivalent candidates from another view $\tau$ as positive samples, denoted by \(\{z_{i^+_1}^\tau, z_{i^+_2}^\tau, z_{i^+_{(\omega-1)}}^\tau\}\).
% For each area task, we randomly select one candidate's prediction \( p_a^i \) from view \( i \) as the query and treat the remaining three semantically equivalent candidates from view \( j \) as positive samples \( \{z_{a,1}^j, z_{a,2}^j, z_{a,3}^j\} \). 
The contrastive loss for area tasks is then defined as:
\begin{equation}
    \mathcal{L}_{\text{area}}(\xi, \tau) = \frac{1}{(\omega-1)n_a}\sum_{i=1}^{n_a} \sum_{j=1}^{\omega-1} \mathcal{D}(q_i^\xi, z_{i^+_j}^\tau).
\end{equation}
Similarly, for each line task, we randomly select one candidate’s prediction \( q_i^\xi \) and treat its directionally opposite candidate from view \( \tau \), denoted \( z_{i^+}^\tau \), as the positive sample. The contrastive loss for line tasks is computed as:
\begin{equation}
    \mathcal{L}_{\text{line}}(\xi, \tau) = \frac{1}{n_l} \sum_{i=1}^{n_l} \mathcal{D}(q_i^\xi, z_{i^+}^\tau).
\end{equation}
Combining all view pairs, the total intra-task contrastive loss is expressed as:
\begin{equation}\label{eq: IT_CL}
\small
    \mathcal{L}_{\text{CL}}^{\text{intra}} = -\frac{1}{2\mathbb{A}(\mathbb{A}-1)} \sum_{\xi=1}^{\mathbb{A}} \sum_{\tau \ne \xi} \left[ \mathcal{L}_{\text{area}}(\xi,\tau) + \mathcal{L}_{\text{line}}(\xi,\tau) \right].
\end{equation}

This loss encourages semantically equivalent candidates of the same task to lie closer in the latent space, enhancing intra-task feature compactness. By incorporating this structured contrastive signal, the encoder becomes more sensitive to the geometric coherence of non-point tasks, an essential property in CGRP, where multiple candidate representations may correspond to the same task.

\noindent\textbf{RL Loss.}
To optimize the CGRP policy, we use the REINFORCE algorithm~\cite{williams1992simple}, a classical Monte Carlo policy gradient method, to train the backbone model. The expected loss is formulated as:
\begin{equation}
    \mathcal{L}_{\text{RL}}^{\theta} = -\mathbb{E}_{g\sim \mathcal{G}} \mathbb{E}_{\pi\sim p_{\theta}(\pi | g)}[\mathcal{R}(\pi)].
\end{equation}
% where $R(\pi)=r_T$ denotes the total reward (i.e., negative cost) of route $\pi$. 
% and $\theta = \{\theta_{\text{enc}}, \theta_{\text{dec}}\}$ are the encoder and decoder parameters.
To reduce variance, we apply a shared baseline $\tilde{\mathcal{R}}(g)$ and approximate the gradient via Monte Carlo sampling. Given a batch of $\mathbb{A}$ augmented views $\{g^\xi\}_{\xi=1}^{\mathbb{A}}$ and $N$ sampled solutions per instance $g$, the empirical gradient is calculated as:
\begin{equation}
\small
    \nabla_{\theta} \mathcal{L}_{\text{RL}}^{\theta} \approx  -\frac{1}{N\mathbb{A}} \sum_{\xi=1}^{\mathbb{A}} \sum_{j=1}^{N} \left(\mathcal{R}(\pi^{\xi,j}) - \tilde{\mathcal{R}}(g)\right) \nabla_{\theta} \log p_{\theta} \left( \pi^{\xi,j}|g^\xi \right),
\end{equation}
with the baseline $\tilde{\mathcal{R}}(g) = \frac{1}{N\mathbb{A}} \sum_{\xi=1}^{\mathbb{A}} \sum_{j=1}^{N} \mathcal{R}(\pi^{\xi,j})$.

\noindent\textbf{Global Objective.}
To enhance the backbone’s robustness and geometry-awareness in solving CGRP, we design a triplet loss incorporating auxiliary contrastive objectives at both the instance and intra-task levels. The instance level CL loss $\mathcal{L}_{\text{CL}}^{\text{ins}}$ promotes consistent global representations across augmented views, while the intra-task CL loss $\mathcal{L}_{\text{CL}}^{\text{intra}}$ encourages semantically coherent representations among candidates belonging to the same task.
The overall training objective is a weighted combination of the reinforcement and contrastive components:
\begin{equation} \label{eq:triplet_loss}
    \mathcal{L}_{\text{total}} = \mathcal{L}_{\text{RL}}^{\theta} + \lambda_{\text{ins}} \mathcal{L}_{\text{CL}}^{\text{ins}} + \lambda_{\text{intra}} \mathcal{L}_{\text{CL}}^{\text{intra}},
\end{equation}
where $\lambda_{\text{ins}}$ and $\lambda_{\text{intra}}$ are tunable hyperparameters that balance the auxiliary losses. The encoder, projection head, and predictor parameters are optimized jointly by the triplet loss, while the decoder is only optimized by the RL loss.
% This joint objective enhances the backbone’s robustness and geometry-awareness in solving CGRP.

\section{Experiment} 
\label{sec:experiments}
\subsection{Experimental Setting}
\noindent\textbf{Training Data.} 
% We conduct extensive experiments to evaluate DiCon on CGRP. 
During training, we use a mixed-composition CGRP distribution. Specifically, the problem size is sampled uniformly from $[20,100)$, the number of area tasks is sampled from $[0,5)$, and the number of line tasks is sampled from $[0,20)$, with the remaining tasks treated as point tasks. All task coordinates are uniformly sampled from the unit square $[0, 1]^2$, subject to geometric constraints that prevent overlap. For training-time parameterization, area tasks use zigzag coverage paths aligned with the longer side to reduce traversal cost, with $\omega=4$ candidate service options; details are provided in Appendix~\ref{sec:zigzag}. We further evaluate irregular area tasks with alternative coverage methods and different values of $\omega$ in Appendix~\ref{sec:real-world}. 
This training distribution mainly contains point-dominant hybrid instances, while single-geometry and more extreme compositions are reserved as held-out settings for evaluating compositional generalization. 

\noindent\textbf{Evaluation Data.} For evaluation, we first consider in-distribution instances with problem sizes $20$ and $100$, where the numbers of line and area tasks follow the same sampling ranges as in training.
We then assess scalability by testing on larger instances with $200$ and $300$ tasks, using the same sampling strategy. 
To further examine compositional generalization across diverse geometry compositions, we construct specialized test sets with different task configurations. \texttt{Point20} and \texttt{Point100} contain only point tasks, with 20 and 100 tasks, respectively. \texttt{Line20} and \texttt{Line100} contain only line tasks, while \texttt{Area20} and \texttt{Area100} contain only area tasks. In addition, \texttt{Area20+Line30} and \texttt{Area20+Line80} mix area and line tasks with moderate and high line densities, respectively. For each problem setting, we generate 200 test instances to form the test dataset.

\noindent\textbf{Hyperparameter.} We configure the neural network with an embedding dimension of 128, 6 encoder layers, and 8 attention heads with a query-key-value dimension of 16 per head. The feed-forward network hidden dimension is set to 512, and logit clipping $C$ is set to 10. For the double-level CL, we generate $\mathcal{A} = 4$ augmented views per instance through rotation and reflection transformations. The weights of the instance-level and intra-task contrastive losses are set to $\lambda_{ins}=0.1$ and $\lambda_{intra}=0.02$, respectively. The hyperparameter study for double-level CL is provided in Appendix~\ref{sec:CL_hyperparam}. The model parameters are optimized using Adam with a learning rate of $1 \times 10^{-4}$ and a weight decay of $1 \times 10^{-6}$. The learning rate is decayed by a factor of 0.1 at epoch 180. The model is trained for 200 epochs with each epoch processing 100,000 instances and a batch size of 64. During inference, we use greedy rollout for evaluation. Methods with the “-Aug” suffix apply instance augmentation~\cite{kwon2020pomo} to further improve solution quality.

\begin{table*}[t]
\centering
\caption{Performance comparison. Left: in-distribution testing. Right: generalization to larger sizes.}
\vspace{-1mm}
\label{tab:cgrp_performance}
\renewcommand\arraystretch{0.97}
\resizebox{\textwidth}{!}{
\begin{tabular}{l|cccccc|cccccc}
\toprule
 & \multicolumn{3}{c}{20} & \multicolumn{3}{c|}{100} & \multicolumn{3}{c}{200} & \multicolumn{3}{c}{300} \\
 & Obj & Gap (\%) & Time & Obj & Gap (\%) & Time & Obj & Gap (\%) & Time & Obj & Gap (\%) & Time \\
\midrule
Gurobi       & \textbf{6.065}\textsuperscript{*} & \textbf{-0.02} & 8.49m  & 8.423 & 0.82 & 2.86h  & 14.451 & 5.66 & 5.03h & 16.253 & 7.17 & 10.08h \\
\midrule
LKH3         & \underline{6.066} & \underline{0.00}  & 7.53m  & \textbf{8.354} & \textbf{0.00} & 50.30m & \textbf{13.677} & \textbf{0.00} & 3.32h & \textbf{15.165} & \textbf{0.00} & 7.96h \\
OR-Tools     & 6.079 & 0.20 & 6.67m & 8.593 & 2.85 & 33.35m & 14.073 & 2.89 & 1.67h & 15.721 & 3.67 & 2.78h \\
ALNS         & 6.127 & 0.99 & 1.03m & 8.525 & 2.05 & 33.57m & 14.466 & 5.77 & 1.69h & 16.642 & 9.74 & 2.83h \\
\midrule
POMO         & 6.076 & 0.15  & 0.09s  & 8.465 & 1.32 & 0.19s  & 14.115 & 3.20 & 0.94s & 16.344 & 7.78 & 2.58s \\
DiCon-P      & 6.075 & 0.14  & 0.07s  & 8.406 & 0.63 & 0.16s  & 13.949 & 1.99 & 0.80s & 16.021 & 5.65 & 2.14s \\
RELD         & 6.074 & 0.13  & 0.09s  & 8.406 & 0.63 & 0.26s  & 13.922 & 1.79 & 1.17s & 15.839 & 4.45 & 2.99s \\
DiCon-R      & 6.072 & 0.10  & 0.08s  & 8.388 & 0.41 & 0.31s  & 13.845 & 1.23 & 1.83s & 15.702 & 3.54 & 4.55s \\
\midrule
POMO-Aug     & \underline{6.066} & \underline{0.00}  & 1.43s  & 8.402 & 0.57 & 2.40s  & 13.978 & 2.20 & 15.07s & 16.129 & 6.36 & 31.88s \\
DiCon-P-Aug  & \underline{6.066} & \underline{0.00}  & 1.42s  & 8.376 & 0.26 & 2.69s  & 13.854 & 1.29 & 17.05s & 15.853 & 4.54 & 38.96s \\
RELD-Aug     & \textbf{6.065} & \textbf{-0.02} & 1.36s  & 8.373 & 0.23 & 3.42s  & 13.824 & 1.07 & 17.65s & 15.724 & 3.69 & 35.67s \\
DiCon-R-Aug  & \textbf{6.065} & \textbf{-0.02} & 1.36s  & \underline{8.363} & \underline{0.10} & 4.01s  & \underline{13.755} & \underline{0.57} & 20.95s & \underline{15.595} & \underline{2.84} & 43.44s \\
\bottomrule
\end{tabular}
}
\vspace{1mm}
\footnotesize
\textsuperscript{*} The result is obtained by Gurobi without a time limit and thus represents the exact optimum.
\vspace{-6mm}
\end{table*}

\noindent\textbf{Baseline.} We compare our approach against several state-of-the-art algorithms, including both traditional and RL-based neural solvers. For traditional solvers, we utilize LKH3~\cite{helsgaun2017lkh3}, a heuristic typically producing near-optimal solutions for TSP variants, Google OR-Tools~\cite{ortools_routing}, an efficient optimization toolkit with specialized routing capabilities, and Adaptive Large Neighborhood Search (ALNS)~\cite{smith2017glns}, a destroy-and-repair metaheuristic. For neural solvers, we compare wth POMO~\cite{kwon2020pomo}, an RL-based method exploiting problem symmetry for routing problems, and RELD~\cite{huang2025rethinking}, which enhances decoder capacity through identity mapping and feed-forward layers to improve generalization. We also compare with Gurobi, a commercial exact solver that provides optimal solutions for size 20 instances. For larger scales, Gurobi is given time limits of $60s$, $90s$, and $180s$ for sizes $100$, $200$, and $300$, respectively. In the compositional generalization experiments, Gurobi provides optimal solutions for \texttt{Point20} and \texttt{Line20} without time limits. For the remaining settings, we impose time limits as follows: 10s for \texttt{Area20}, 20s for \texttt{Area20+Line30}, and 60s for \texttt{Point100}, \texttt{Line100}, \texttt{Area100}, and \texttt{Area20+Line80}. We apply DiCon to two neural routing backbones, POMO and RELD, resulting in DiCon-P and DiCon-R, respectively. All methods are executed on a machine equipped with an NVIDIA A100-SXM4-80GB GPU and an AMD EPYC 7742 64-Core Processor. The code and dataset are publicly released for reproducibility.\footnote{https://github.com/mingfan321/Learning-to-Solve-Compositional-Geometry-Routing-Problems}

\noindent\textbf{Metric.} We evaluate algorithm performance using three metrics: the average objective value (Obj.) representing the total tour length; the average gap (Gap) calculated as (Obj - Baseline)/Baseline × 100\%; and total run time (Time) per test dataset measured in CPU seconds. We select LKH3 as the primary Baseline for computing gaps due to its strong and consistent performance among solvers. We highlight the best Obj and Gap in \textbf{bold} and \underline{underline} the second-best Obj and Gap.

\subsection{Comparison Analysis}
% The comparison results are reported in Table~\ref{tab:cgrp_performance} including in-distribution instances and larger-size generalization settings. DiCon-R consistently outperforms RELD across all test sets, with the performance gap widening as problem size increases.
% Similarly, DiCon-P surpasses POMO in every evaluated setting. With data augmentation, DiCon’s performance improves further. On size-20 instances, DiCon-R-Aug matches the Gurobi optimum, while both DiCon-R-Aug and DiCon-P-Aug surpass Gurobi on size-50 and 100 instances.
% In terms of efficiency, DiCon variants significantly reduce runtime compared to exact and traditional heuristic solvers, while maintaining comparable time with POMO and RELD.

The results are reported in Table~\ref{tab:cgrp_performance}, covering both in-distribution instances and larger-size generalization settings. DiCon consistently improves different neural routing backbones. On in-distribution instances, DiCon-P and DiCon-R achieve competitive solution quality with substantially lower runtime than Gurobi and LKH3. On larger 200-task and 300-task instances, DiCon-P and DiCon-R clearly outperform their original backbones, POMO and RELD, with the performance gap becoming more pronounced as the problem size increases. With data augmentation, DiCon achieves further improvements in solution quality. In terms of efficiency, DiCon variants require much less runtime than exact and traditional heuristic solvers. These results demonstrate that DiCon provides a strong quality-efficiency trade-off and robust generalization to problem sizes beyond the training range.

\begin{table*}[t]
\centering
\caption{Performance comparison across different compositional geometry instances.}
\vspace{-1mm}
\label{tab:results_generalization}
\renewcommand\arraystretch{0.97}  % 0.97
\resizebox{\textwidth}{!}{
\begin{tabular}{l|ccc|ccc|ccc|ccc}
\toprule
& \multicolumn{3}{c|}{\texttt{Point20} (TSP)} & \multicolumn{3}{c|}{\texttt{Point100} (TSP)} & \multicolumn{3}{c|}{\texttt{Line20}} & \multicolumn{3}{c}{\texttt{Line100}} \\
     & Obj & Gap (\%) & Time & Obj & Gap (\%) & Time & Obj & Gap (\%) & Time & Obj & Gap (\%) & Time \\
\midrule
Gurobi      & \textbf{3.918}\textsuperscript{*} & \textbf{0.00}  & 5.27m  & 8.011 & 2.53 & 3.23h  & \textbf{4.339}\textsuperscript{*} & \textbf{-0.02} & 7.10h  & \textbf{12.184} & \textbf{-3.17} & 2.82h \\
\midrule
LKH3        & \textbf{3.918} & \textbf{0.00}  & 4.93m  & \textbf{7.813} & \textbf{0.00} & 20.98m & \underline{4.340} & \underline{0.00}  & 8.57m  & 12.582 & 0.00  & 29.77m \\
OR-Tools    & 3.918 & 0.00  & 6.68m    & 8.006 & 2.47 & 33.34m & 4.356 & 0.36 & 6.68m & 13.398 & 6.48 & 33.38m \\
ALNS        & 3.930 & 0.30  & 33.03s   & 7.911 & 1.25 & 44.30m & 4.382 & 0.97 & 1.08m & 12.697 & 0.91 & 43.61m \\
\midrule
POMO        & 3.923 & 0.11  & 0.03s  & 7.903 & 1.16 & 0.21s  & 4.347 & 0.16  & 0.03s  & 12.623 & 0.32  & 0.71s \\
DiCon-P     & 3.921 & 0.06  & 0.02s  & 7.857 & 0.56 & 0.25s  & 4.343 & 0.07  & 0.04s  & 12.487 & -0.76 & 0.76s \\
RELD        & 3.920 & 0.05  & 0.03s  & 7.855 & 0.55 & 0.25s  & 4.345 & 0.12  & 0.03s  & 12.471 & -0.88 & 0.68s \\
DiCon-R     & 3.920 & 0.04  & 0.03s  & 7.844 & 0.40 & 0.33s  & 4.341 & 0.02  & 0.05s  & 12.394 & -1.50 & 0.92s \\
\midrule
POMO-Aug    & \underline{3.919} & \underline{0.00}  & 0.20s  & 7.850 & 0.47 & 1.61s  & \underline{4.340} & \underline{0.00}  & 0.26s  & 12.494 & -0.71 & 4.36s \\
DiCon-P-Aug & 3.919 & 0.01  & 0.19s  & 7.831 & 0.24 & 1.95s  & \textbf{4.339} & \textbf{-0.02} & 0.27s  & 12.389 & -1.54 & 6.05s \\
RELD-Aug    & \textbf{3.918} & \textbf{0.00}  & 0.22s  & 7.829 & 0.21 & 2.03s  & \underline{4.340} & \underline{0.00}  & 0.26s  & 12.382 & -1.59 & 5.17s \\
DiCon-R-Aug & \textbf{3.918} & \textbf{0.00}  & 0.22s  & \underline{7.822} & \underline{0.12} & 2.61s  & \textbf{4.339} & \textbf{-0.02} & 0.31s  & \underline{12.319} & \underline{-2.10} & 7.71s \\
\midrule
\midrule
& \multicolumn{3}{c|}{\texttt{Area20}} & \multicolumn{3}{c|}{\texttt{Area100}} & \multicolumn{3}{c|}{\texttt{Area20+Line30}} & \multicolumn{3}{c}{\texttt{Area20+Line80}} \\
     & Obj & Gap (\%) & Time & Obj & Gap (\%) & Time & Obj & Gap (\%) & Time & Obj & Gap (\%) & Time \\
\midrule
Gurobi      & \textbf{9.387} & \textbf{-0.26} & 22.2m  & 24.250 & 0.20 & 3.33h  & \underline{12.394} & \underline{-1.04} & 1.02h & \textbf{16.078} & \textbf{-2.62} & 1.86h \\
\midrule
LKH3        & 9.411 & 0.00  & 8.17m  & \textbf{24.200} & \textbf{0.00} & 27.67m & 12.524 & 0.00  & 16.00m & 16.510 & 0.00  & 43.63m \\
OR-Tools    & 9.443 & 0.35 & 6.70m & 24.841 & 2.65 & 33.55m & 12.637 & 0.91 & 16.71m & 16.791 & 1.70 & 33.41m \\
ALNS        & 9.523 & 1.20 & 1.07m & 24.359 & 0.66 & 43.16m & 12.665 & 1.12 & 11.85m & 16.628 & 0.71 & 33.82m \\
\midrule
POMO        & 9.469 & 0.62  & 0.04s  & 24.725 & 2.17 & 2.57s  & 12.504 & -0.16 & 0.16s  & 16.516 & 0.03  & 0.53s \\
DiCon-P     & 9.459 & 0.52  & 0.06s  & 24.483 & 1.17 & 3.39s  & 12.470 & -0.43 & 0.23s  & 16.390 & -0.73 & 0.80s \\
RELD        & 9.466 & 0.59  & 0.05s  & 24.472 & 1.12 & 3.11s  & 12.461 & -0.50 & 0.19s  & 16.370 & -0.85 & 1.01s \\
DiCon-R     & 9.436 & 0.27  & 0.07s  & 24.303 & 0.42 & 3.41s  & 12.437 & -0.70 & 0.27s  & 16.303 & -1.25 & 1.37s \\
\midrule
POMO-Aug    & 9.417 & 0.06  & 0.44s  & 24.543 & 1.42 & 14.78s & 12.426 & -0.78 & 2.35s  & 16.372 & -0.84 & 7.45s \\
DiCon-P-Aug & 9.411 & 0.00  & 0.45s  & 24.369 & 0.70 & 20.19s & 12.408 & -0.93 & 3.57s  & 16.287 & -1.35 & 8.39s \\
RELD-Aug    & 9.419 & 0.09  & 0.51s  & 24.354 & 0.63 & 17.04s & 12.402 & -0.98 & 2.41s  & 16.277 & -1.41 & 8.97s \\
DiCon-R-Aug & \underline{9.400} & \underline{-0.11} & 0.57s  & \underline{24.217} & \underline{0.07} & 23.71s & \textbf{12.385} & \textbf{-1.11} & 3.66s  & \underline{16.217} & \underline{-1.78} & 10.87s \\
\bottomrule
\end{tabular}
}
\vspace{1mm}
\footnotesize
\noindent\textsuperscript{*} The result is obtained by Gurobi without a time limit and thus represents the exact optimum.
\vspace{-3mm}
\end{table*}

\subsection{Performance Across Diverse Geometry Compositions}

We consider two types of geometry composition settings: single-geometry and extreme hybrid-geometry compositions. The experimental results are reported in Table~\ref{tab:results_generalization}.

\noindent\textbf{Performance on Single-Geometry Settings.}
DiCon achieves consistent improvements across single-geometry settings, with the advantage becoming more pronounced as the problem size increases. On \texttt{Point20} and \texttt{Line20}, DiCon-R-Aug reaches optimal solution quality. Notably, line-only and area-only instances are not included in the training distribution, suggesting that DiCon can generalize to held-out geometry compositions. DiCon-R-Aug remains competitive with LKH3 while requiring much less runtime, and even achieves better solution quality on \texttt{Line20}, \texttt{Line100}, and \texttt{Area20}.

\noindent\textbf{Performance on Extreme hybrid-geometry.}
For unseen extreme hybrid compositions, DiCon-R-Aug achieves best performance on \texttt{Area20+Line30} and outperforms LKH3 on \texttt{Area20+Line30} and \texttt{Area20+Line80}, suggesting that DiCon is effective in handling extreme hybrid-geometry. 

\begin{wrapfigure}{r}{0.35\textwidth}
    \centering
    \vspace{-12mm}
    \includegraphics[width=\linewidth]{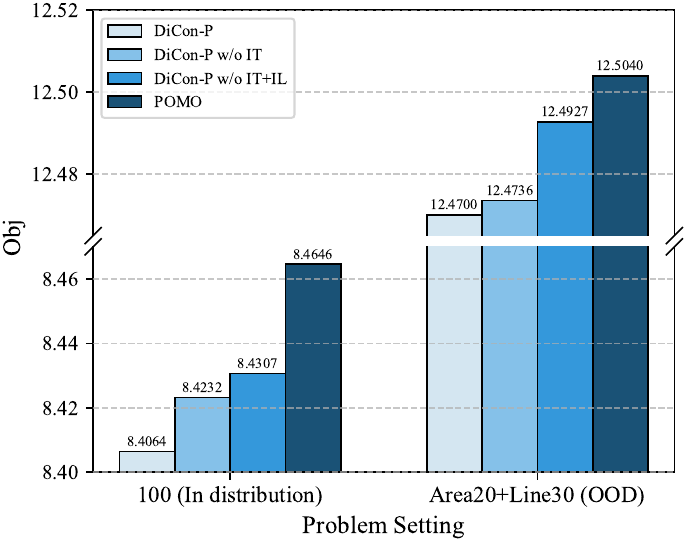}
    \vspace{-5mm}
    \caption{Ablation study.}
    \label{fig:ablation}
    \vspace{-12mm}
\end{wrapfigure}

\subsection{Ablation Study}
To assess the contribution of key components in DiCon, we progressively remove the intra-task (IT) CL, instance-level (IL) CL, and DA modules from DiCon-P, yielding DiCon-P w/o IT, DiCon-P w/o IT+IL, and the backbone POMO, respectively. The results on CGRP100 and \texttt{Area20+Line30} are visualized in Fig.~\ref{fig:ablation}. Performance consistently degrades as components are removed, demonstrating the complementary and positive contributions of each module.

\section{Conclusion}
\label{sec:conclu}
In this paper, we study the Compositional Geometry Routing Problem (CGRP), which generalizes conventional routing by covering point, line, area, and hybrid task geometries that commonly arise in real-world applications. We formulate CGRP solving as an MDP and identify its key characteristics and challenges. 
Our primary contribution is DiCon, a plug-and-play framework that combines a differential attention mechanism with a double-level contrastive objective to enhance both representation learning and decision-making. Empirically, DiCon consistently demonstrates strong performance, broad versatility, and robust compositional generalization across diverse problem settings and backbone models. The present formulation focuses on single-agent routing without complex operational constraints, such as time windows and capacity limits. Future work will extend DiCon with constraint-aware modules and investigate multi-agent CGRP, where hierarchical or decentralized RL may be required for effective coordination.

\begin{ack}
This research is partly supported by the National Research Foundation, Singapore under its AI Singapore Programme (AISG Award No: AISG3-RP-2022-031).
\end{ack}

{
\small
\bibliographystyle{plain}
\bibliography{ref}
}

%%%%%%%%%%%%%%%%%%%%%%%%%%%%%%%%%%%%%%%%%%%%%%%%%%%%%%%%%%%%
\newpage
\appendix
\section{DiCon Backbones}
\label{appendix_model}

Most neural routing solvers adopt a transformer encoder-decoder architecture. Given an instance \(g \in \mathcal{G}\), the encoder produces node embeddings that represent the instance, and the decoder then iteratively selects actions conditioned on these embeddings. To demonstrate the plug-and-play nature of DiCon, we integrate it into two strong neural solvers, POMO~\cite{kwon2020pomo} and RELD~\cite{huang2025rethinking}, resulting in \emph{DiCon-P} and \emph{DiCon-R}, respectively.

\subsection{POMO}
Given an instance \(g\in \mathcal{G}\), we first obtain initial embeddings \(H^0=\{h_0^0,\ldots,h_{|\mathcal{V}|}^0\}\) using the unified encoding scheme in Section~\ref{sec: method_overview}. These embeddings are then processed by \(\mathbb{L}\) encoder layers to yield the final embeddings \(H^{\mathbb{L}}=\{h_0^{\mathbb{L}},\ldots,h_{|\mathcal{V}|}^{\mathbb{L}}\}\). Each encoder layer consists of an MHA sublayer, a residual Add \& Norm (skip connection with instance normalization), a feed-forward (FF) sublayer, and a second Add \& Norm.

Specifically, at layer \(l\in\{1,\ldots,\mathbb{L}\}\), the embeddings are updated by 8-head MHA followed by Add \& Norm:
\begin{equation}
\hat h_i
= \mathrm{IN}\bigl(h_i^{l-1}
  + \mathrm{MHA}\bigl(h_0^{l-1},\dots,h_{|\mathcal{V}|}^{l-1}\bigr)\bigr),
\quad \forall\,i \in\{0,\dots,|\mathcal{V}|\}.
\label{ap_eq:W-node}
\end{equation}
Then, an FF sublayer and another Add \& Norm produce \(\{h_i^{l}\}_{i=0}^{|\mathcal{V}|}\):
\begin{equation}
h_i^{l} = \mathrm{IN}\bigl(\hat h_i + \mathrm{FF}(\hat h_i)\bigr).
\label{ap_eq:w-ff}
\end{equation}

Given the final embeddings \(\{h_i^{\mathbb{L}}\}_{i=0}^{|\mathcal{V}|}\), the decoder autoregressively computes the candidate-selection distribution over \(T\) steps. At decoding step \(t\in\{1,\ldots,T\}\), the context embedding \(h_c\) is produced by an 8-head MHDA layer based on contextual features \(v_c=[h_{\pi_1},\,h_{\pi_{t-1}}]\) and the encoder embeddings:
\begin{equation}
h_c \;=\; \mathrm{MHDA}\Bigl(v_c,\;\{\,h_0^\mathbb{L},\dots,h_{|\mathcal{V}|}^\mathbb{L}\}\Bigr).
\label{ap_eq:decoder_mha}
\end{equation}
The context embedding \(h_c\) is then used to compute unnormalized compatibility scores \(\alpha\):
\begin{equation}
\alpha_i =
\begin{cases}
-\infty, & \text{if candidate }i\text{ is masked},\\[2pt]
C\cdot\tanh\!\Bigl(\tfrac{h_c^\top (W^K h^\mathbb{L}_i)}{\sqrt{d}}\Bigr), & \text{otherwise},
\end{cases}
\label{ap_eq:decoder_com}
\end{equation}
where \(C=10\) and \(W^K\) is a learnable projection. Finally, the selection probability is
\begin{equation}
p_\theta(\pi_t \mid \pi_{<t}, g) \;=\; \operatorname{Softmax}(\alpha).
\label{ap_eq:softmax}
\end{equation}

POMO employs a multi-trajectory strategy during both training and inference by sampling multiple decoding trajectories for each instance, each initialized from a different start node. During training, the baseline is set to the mean reward (negative tour length) across sampled trajectories. At inference, trajectories are generated in parallel and the one with the minimum cost is selected, improving solution-space exploration with modest overhead.

\subsection{RELD}
RELD follows the same encoder as POMO but modifies the decoder to inject context information directly into the embedding space, rather than relying solely on attention weights. Concretely, it adds a residual connection from the most recently visited node \(h_{\pi_{t-1}}\) to the context embedding:
\begin{equation}
h_c = \mathrm{MHDA}(v_c, \{h_0^\mathbb{L},\dots,h_{|\mathcal{V}|}^\mathbb{L}\}) + h_{\pi_{t-1}}.
\end{equation}
In addition, RELD augments the decoder with a lightweight feed-forward network and a residual connection:
\begin{equation}
q_c = h_c + \mathrm{FF}(h_c) = h_c + \sigma\!\left(W_2\,\sigma\!\left(W_1 h_c + b_1\right) + b_2\right),
\end{equation}
where \(W_1, W_2, b_1,\) and \(b_2\) are learnable parameters and \(\sigma\) denotes the ReLU activation. Together, these modifications form a transformer-style update with a single query, enabling nonlinear processing of the query representation and allowing richer context transformations within the decoder.

\section{Training Algorithm}
The training process is outlined in Algorithm~\ref{alg: triplet_loss}. For each training iteration, we sample a batch of base instances $\{g_b\}_{b=1}^{B}$ (Line 3) and generate $\mathbb{A}$ augmented views for each, resulting in $\mathbb{A} \cdot B$ instances in total (Line 4). The model then computes $N$ solutions for each view $g_b^\xi$ using the learned policy $p_{\theta}(\pi^{b,\xi,j} | g_b^\xi)$ (Line 7), where $\xi \in \{1, \ldots, \mathbb{A}\}$ and $j \in \{1, \ldots, N\}$. Finally, model parameters are updated jointly using the triplet loss derived from the contrastive objectives and reinforcement loss (Lines 8-10).

\label{appendix:alg}
\begin{algorithm}[!t]
    \caption{Training Algorithm for DiCon}
    \label{alg: triplet_loss}
    \textbf{Input}: Instance distribution $\mathcal{G}$, number of augmentations $\mathbb{A}$, number of training steps $E$, batch size $B$, number of tours $N$ per subproblem;\\
    \textbf{Output}: The trained policy network $\theta$;
    
    \begin{algorithmic}[1] %[1] enables line numbers
        % \STATE Let $t=0$.
        \STATE Initialize policy network $\theta$.
        \FOR{$e = 1$ to $E$}
        
        \STATE{$g_b \sim $ \textsc{SampleInstance} ($\mathcal{G}$),\quad $\forall b \in \left\{1, \cdots, B\right\}$}

        \STATE{$\{g_b^1, \cdots, g_b^\mathbb{A}\} \sim $ \textsc{DataAugmentation} ($g_b$), \quad $\forall b \in \left\{1, \cdots, B\right\}$}

        \STATE{$\{z_b^1, \cdots, z_b^\mathbb{A}\} \sim $ \textsc{ProjectionHead} ($g_b$), \quad $\forall b \in \left\{1, \cdots, B\right\}$}

        \STATE{$q_b^\xi \sim $ \textsc{PredictionHead} ($z_b^\xi$), \quad $\forall \xi \in\{1, \cdots, \mathbb{A}\}, \quad \forall b \in \left\{1, \cdots, B\right\}$}
        
        \STATE{$\pi^{b,\xi,j}\! \sim $\! \textsc{SampleSolutions} ($p_{\theta}(\cdot|g_b^\xi)$), $\forall j \in \left\{1, \cdots, N\right\},\quad \forall \xi \in\{1, \cdots, \mathbb{A}\}, \quad \forall b \in \left\{1, \cdots, B\right\}$}
        
        \STATE{Calculate instance-level CL loss and intra-level CL loss according to Eq.~(\ref{eq: IL_CL}) and Eq.~(\ref{eq: IT_CL})} 
        
        \STATE{Calculate gradient $\nabla_{\theta} \mathcal{L}_{total}$ according to Eq.~(\ref{eq:triplet_loss})}
        
        \STATE{$\theta \leftarrow$ \text{ADAM}($\theta, \nabla_{\theta} \mathcal{L}_{total}$)}
        \ENDFOR
    \end{algorithmic}
\end{algorithm}

\section{Zigzag Path Generation for Area Tasks}
\label{sec:zigzag}

For training-time parameterization, we define a zigzag scanning path within each area task region to simulate a sensor-sweeping pattern, such as a UAV camera or ultrasonic probe. The length of this path is used as the coverage cost in the CGRP instance.

\textbf{Rotated Rectangle Construction.} 
Each area task is represented as a rotated rectangle with center anchor, length $\tilde{L}$, width $\tilde{W}$, and orientation $\beta$. The length $\tilde{L}$ is set as the minimal anchor distance for area tasks, while the width $\tilde{W}$ is sampled from a uniform range $[3\gamma, \tilde{L})$. The rectangle corners are computed by rotating the axis-aligned corners via the matrix $\tilde{R}(\beta)$ and translating them to the anchor point.

\textbf{Zigzag Path Length Estimation.}
To perform coverage, we align the sweeping path along the rectangle’s longer side. The number of parallel sweeps is computed as:
\begin{equation}
n_{\text{sweep}} = \left\lfloor \frac{\tilde{W}}{\gamma} \right\rfloor + 1,
\end{equation}
and the total zigzag path length is estimated as:
\begin{equation}
L_{\text{zigzag}} = n_{\text{sweep}} \cdot \tilde{L}.
\end{equation}
\textbf{Entry–Exit Pair Sampling.}
For each area task, we define 4 entry-exit pairs associated with the two short edges of the rectangular region.
Specifically, entry points are sampled at the endpoints of each short edge.
Given the parity of $n_{\text{sweep}}$, the exit point is offset from the entry either on the same side or the opposite side, following the zigzag path displacement.

% For each area task, we define four entry–exit pairs located at the short edges of the rectangle. Entry points are sampled on one short edge, and their corresponding exit points are projected onto the opposite long edge, offset by the total sweep displacement. The displacement direction depends on the parity of $n_{\text{sweep}}$, ensuring consistent path traversal.

\section{Inter-task Level Contrastive Learning}

We further investigate whether inter-task-level contrastive learning (CL) can help models better distinguish between heterogeneous task types. Specifically, tasks of the same type (e.g., all areas) are treated as positive pairs, while tasks of different types serve as negatives. We implement this strategy on top of POMO, yielding a variant termed POMO\_inter\_task\_CL. Results on CGRP instances with 20–100 tasks are reported in Table~\ref{tab: pomo_cl_sizes_small}. Empirically, inter-task CL brings no gains on CGRP-20 and -50, and degrades performance on CGRP-100. We hypothesize that this is due to over-clustering: forcing all latent representations of the same type together may hinder the model’s ability to discriminate fine-grained variations within each task class.

\begin{table*}[t]
\vspace{2mm}
\centering
\caption{Results under different problem sizes.}
\label{tab: pomo_cl_sizes_small}
\renewcommand{\arraystretch}{1.1}
\begin{tabular}{l|ccc}
\hline
Method & 20 & 50 & 100 \\
\hline
POMO & 6.08 & 6.23 & 8.46 \\
POMO\_inter\_task\_CL & 6.08 & 6.23 & 8.48 \\
\hline
\end{tabular}
\vspace{-2mm}
\end{table*}

\section{Implementation Details of the Code}
\label{sec:implementation}

% For a better understanding of the details of DiCon, we give implementation details of the DA and double-level CL components.
To facilitate a deeper understanding of DiCon, we provide the implementation details of its core components, namely the DA mechanism and the double-level CL objectives.

\subsection{Differential Attention}

$\lambda$ controls the strength of the subtractive branch in DA. The attention output is computed as $s_m = \left(A_{m}^{1} - \lambda_m A_{m}^{2}\right)V$, so a larger $\lambda_m$ means stronger suppression from the second attention map, while a smaller $\lambda_m$ makes DA closer to standard attention. The purpose is therefore not to create a new attention map arbitrarily, but to modulate how aggressively potentially low-saliency candidates are discounted. This is consistent with the paper's description of DA as an attention-denoising mechanism for suppressing probability mass on strategically weak actions.

In our implementation, $\lambda_m$ is constrained to be non-negative. Therefore, it cannot become negative during training. It is initialized to $0.5$ and remains in a moderate positive range throughout optimization. Based on the logged values, the final trained values at epoch 200 are $[1.0424,\ 0.3838,\ 0.2412,\ 0.3726,\ 0.2228,\ 0.4419,\ 0.3850,\ 0.4772]$, with a mean of $0.4459$. This indicates that the model does not drive $\lambda_m$ to an extreme regime; rather, it learns head-dependent suppression strengths, with most heads staying below $0.5$ and one head learning substantially stronger subtraction.

\begin{pycode}
import torch
import torch.nn.functional as F

def differential_multi_head_attention(
    q_a: torch.Tensor,
    k_a: torch.Tensor,
    q_b: torch.Tensor,
    k_b: torch.Tensor,
    v: torch.Tensor,
    ninf_mask: torch.Tensor,
    lambda_broadcast: torch.Tensor,
    key_dim: int,
):
    """
    q_a, q_b: (B,H,n,D)
    k_a, k_b: (B,H,problem,D)
    v:        (B,H,problem,D)
    ninf_mask:(B,n,problem)  (here n = pomo)
    lambda_broadcast: (1,H,1,1) or (1,1,1,1)
    returns: (B,n,H*D)
    """
    B, H, n, D = q_a.shape
    problem = k_a.size(2)

    # logits: (B,H,n,problem)
    logits_a = torch.matmul(q_a, k_a.transpose(2, 3)) / math.sqrt(key_dim)
    logits_b = torch.matmul(q_b, k_b.transpose(2, 3)) / math.sqrt(key_dim)

    # apply mask to both branches
    # ninf_mask: (B,n,problem) -> (B,1,n,problem)
    mask = ninf_mask[:, None, :, :].expand(B, H, n, problem)
    logits_a = logits_a + mask
    logits_b = logits_b + mask

    w_a = F.softmax(logits_a, dim=3)  # (B,H,n,problem)
    w_b = F.softmax(logits_b, dim=3)  # (B,H,n,problem)

    weights = w_a - lambda_broadcast * w_b  # (B,H,n,problem)
    out = torch.matmul(weights, v)  # (B,H,n,D)
    out = out.transpose(1, 2).contiguous().view(B, n, H * D)  # (B,n,H*D)
    return out
\end{pycode}

\subsection{Contrastive Learning}

\begin{pycode}
z, q = self.model.pre_forward(reset_state)
z = z.reshape(N_aug, -1, z.shape[1], z.shape[2])
q = q.reshape(N_aug, -1, q.shape[1], q.shape[2])

z = F.normalize(z, dim=-1)
q = F.normalize(q, dim=-1)

sim_loss = 0.0
struct_loss = 0.0
cnt = 0
for i in range(N_aug):
    for j in range(N_aug):
        if i == j:
            continue
        q_view_i = q[i]  # (B, N+1, D) predictor output
        z_view_j = z[j]  # (B, N+1, D) projector output

        # Instance-level contrastive loss
        sim_loss += -(q_view_i * z_view_j.detach()).sum(dim=-1).mean()

        # Intra-task-level contrastive loss
        struct_loss += self.task_structure_loss(
            q_view_i, n_areas, n_lines, target_output=z_view_j
        )
        cnt += 1

loss_siam = sim_loss / cnt
loss_struct = struct_loss / cnt
\end{pycode}

\section{Real-World Scenarios}
\label{sec:real-world}

\begin{figure*}
    \centering\includegraphics[width=0.99\columnwidth]{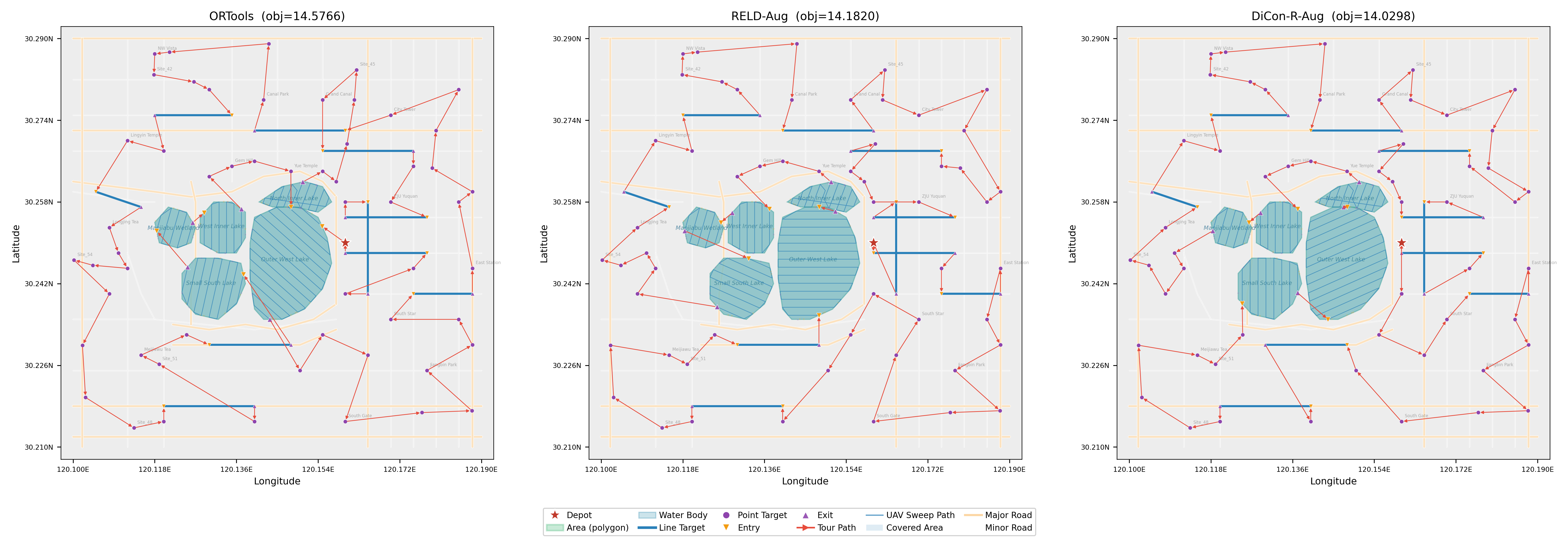}
    \caption{Visualization results on a real-world urban monitoring instance in Hangzhou, where area tasks represent water bodies (e.g., lakes), line tasks correspond to road networks, and point tasks denote key landmarks or facilities requiring localized inspection. 
    }
    \label{fig:urban}
\end{figure*}

\begin{figure*}
    \centering\includegraphics[width=0.99\columnwidth]{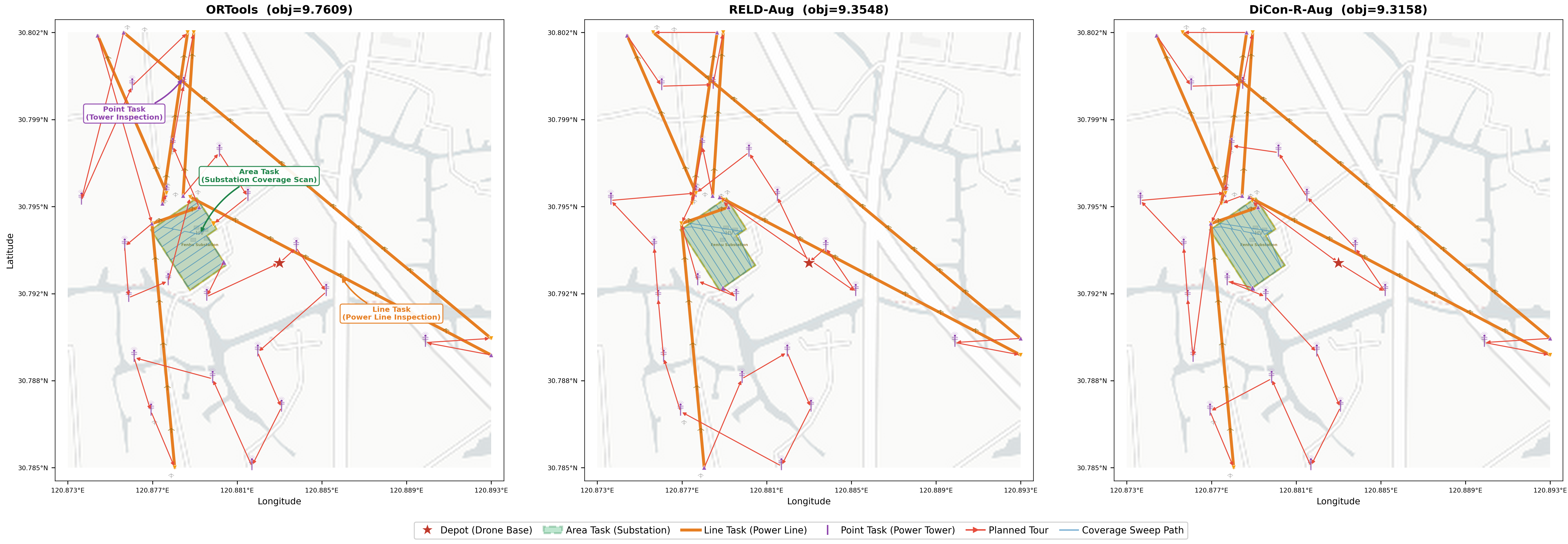}
    \caption{Visualization results on a real-world electric power inspection instance in Jiaxing, where the area task represents the Fenhu substation (approximately 337 m × 400 m), the line tasks correspond to seven transmission lines, and the point tasks represent 20 high-voltage transmission towers.
    }
    \label{fig:power}
\end{figure*}

We include a \emph{heterogeneous urban monitoring scenario}, shown in Fig.~\ref{fig:urban}, where area tasks correspond to water bodies requiring surface inspection, line tasks represent road networks for infrastructure monitoring, and point tasks denote critical facilities or landmarks requiring localized inspection. All locations were obtained from a real-world map of Hangzhou, China. The evaluation is conducted in a zero-shot manner, the area tasks are irregular polygons rather than rectangles, and the coverage path is generated by a coverage path planning (CPP) routine~\cite{bahnemann2021revisiting}. The candidate set is expanded to include the eight shortest entry–exit pairs (i.e., $\omega=8$) induced by the CPP path. 
In addition, we include a real-world experiment on \emph{electric power line inspection}, and the results are shown in Fig.~\ref{fig:power}. We selected the area surrounding the Fenhu 500 kV / 220 kV Substation in Jiaxing, China, where we constructed a real-world CGRP instance. All geographic data were obtained from real annotations in OpenStreetMap.

On both real-world instances, DiCon-R achieves better performance than its backbone model, RELD, and also outperforms OR-Tools. Notably, this test setting differs from the synthetic training setup in region shape, coverage pattern, and candidate cardinality. The two instances therefore exhibit substantial distribution shifts from the training data, further demonstrating the practical applicability and robustness of DiCon.

\section{Visualization Results}
\begin{figure*}
    \centering\includegraphics[width=0.99\columnwidth]{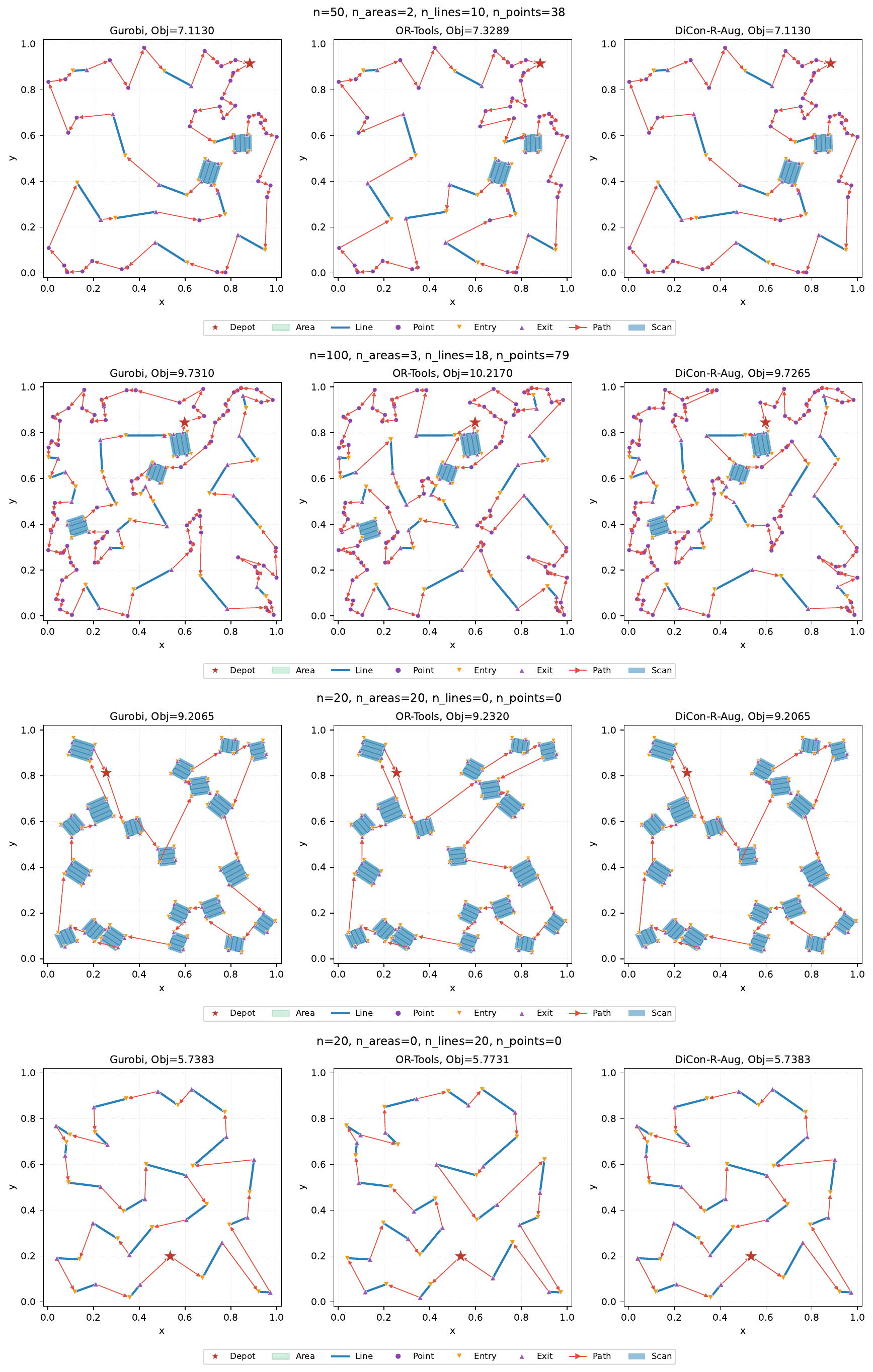}
    \caption{The visualizations of randomly selected CGRP instances.
    }
    \label{fig:routings}
\end{figure*}

We provide qualitative visualizations of randomly selected CGRP instances in Fig.~\ref{fig:routings} to highlight the effectiveness of DiCon in handling diverse task geometries.

\section{Performance on Large-Size Instances}

\begin{figure*}
    \centering\includegraphics[width=0.6\columnwidth]{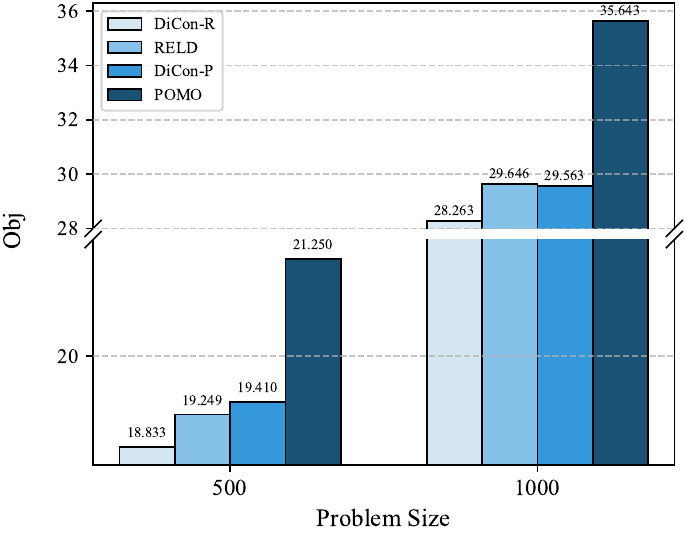}
    \caption{Comparative results on large problem sizes.
    }
    \label{fig:large_size}
\end{figure*}

Figure~\ref{fig:large_size} shows the results on large-size instances, with $20$ test instances for each setting. DiCon consistently improves both backbone models. On 500-task instances, DiCon-P reduces the objective of POMO from 21.250 to 19.410, while DiCon-R improves RELD from 19.249 to 18.833. On 1000-task instances, the gains are more substantial: DiCon-P reduces the objective of POMO from 35.643 to 29.563, and DiCon-R improves RELD from 29.646 to 28.263. These results suggest that DiCon generalizes effectively beyond the training range and provides stronger scalability on large CGRP instances.

\section{Results on Imbalanced CGRP}

\begin{figure*}
    \centering\includegraphics[width=0.99\columnwidth]{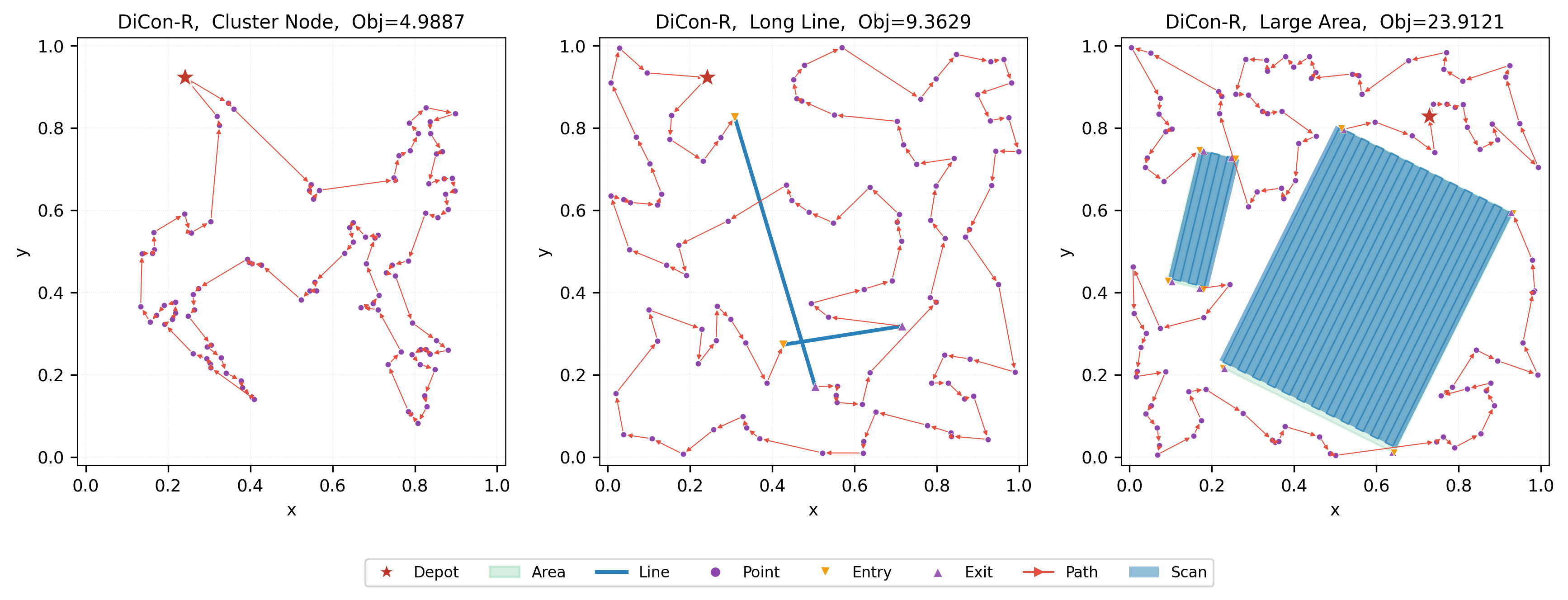}
    \caption{The visualization results on extreme and highly imbalanced task mixtures. (1) Clustered-node instances, where point tasks follow a clustered distribution that differs substantially from the training distribution; (2) Long-line instances, which contain two significantly longer line tasks; and (3) Large-area instances, which contain two substantially larger area tasks.
    }
    \label{fig:imbalanced}
\end{figure*}

We further assess DiCon on three types of extreme instances: clustered-node instances, long-line instances, and large-area instances. The results, reported in Table~\ref{tab:generalization_extreme_task_mixtures} and visualized in Fig.~\ref{fig:imbalanced}, show that DiCon generally maintains a performance advantage under these challenging geometric patterns.

\begin{table*}[t]
\centering
\caption{Generalization results on extreme and highly imbalanced task mixtures.}
\label{tab:generalization_extreme_task_mixtures}
\resizebox{0.6\linewidth}{!}{
\begin{tabular}{lcccccc}
\toprule
\multirow{2}{*}{Algorithm} & \multicolumn{2}{c}{Cluster Node} & \multicolumn{2}{c}{Long Line} & \multicolumn{2}{c}{Large Area} \\
\cmidrule(lr){2-3} \cmidrule(lr){4-5} \cmidrule(lr){6-7}
 & Obj & Time (s) & Obj & Time (s) & Obj & Time (s) \\
\midrule
OR-Tools & 4.9525 & 9.39 & 8.8544 & 30.35 & 20.9451 & 33.15 \\
RELD     & 4.9091 & 0.72 & 8.6683 & 0.55  & 20.8315 & 0.56  \\
DiCon-R  & 4.8927 & 0.52 & 8.6503 & 0.55  & 20.8038 & 0.55  \\
\bottomrule
\end{tabular}
}
\end{table*}

\section{Performance on Generalized TSP}

To further evaluate the applicability of DiCon beyond CGRP, we conduct experiments on Generalized TSP (GTSP), where each cluster contains multiple candidate nodes and the solver must select one node from each cluster while optimizing the tour. The results are reported in Table~\ref{tab:gtsp_results}, with each setting evaluated on 200 instances. DiCon-R achieves the best objective values on GTSP50 and GTSP100, with objectives of $1.6494$ and $2.1174$, respectively. On GTSP20, DiCon-R also remains highly competitive, achieving an objective of $1.1422$, close to the Gurobi result of $1.1372$. Compared with RELD, DiCon-R consistently improves solution quality across all tested sizes. Meanwhile, DiCon-R maintains fast inference, requiring only $0.09$s, $0.10$s, and $0.15$s for GTSP20, GTSP50, and GTSP100, respectively. These results indicate that DiCon-R can generalize effectively to GTSP and provides a favorable balance between solution quality and computational efficiency.

\begin{table*}[t]
\centering
\caption{Experimental results on Generalized TSP.}
\label{tab:gtsp_results}
\resizebox{0.6\linewidth}{!}{
\begin{tabular}{lcccccc}
\toprule
\multirow{2}{*}{Method} & \multicolumn{2}{c}{GTSP20} & \multicolumn{2}{c}{GTSP50} & \multicolumn{2}{c}{GTSP100} \\
\cmidrule(lr){2-3} \cmidrule(lr){4-5} \cmidrule(lr){6-7}
 & Obj & Time & Obj & Time & Obj & Time \\
\midrule
Gurobi   & 1.1372 & 49.42s & 1.6500 & 8.11h  & 2.3147 & 33.33h \\
OR-tools & 1.1427 & 0.60s  & 1.7403 & 4.00s  & 2.4722 & 23.80s \\
ALNS     & 1.2026 & 44.79s & 1.8733 & 1.03m  & 2.5760 & 2.41m  \\
RELD     & 1.1433 & 0.08s  & 1.6828 & 0.08s  & 2.1806 & 0.13s  \\
DiCon-R  & 1.1422 & 0.09s  & 1.6494 & 0.10s  & 2.1174 & 0.15s  \\
\bottomrule
\end{tabular}
}
\end{table*}

\section{Performance on Distance Constrained CGRP}

\begin{figure*}
    \centering\includegraphics[width=0.99\columnwidth]{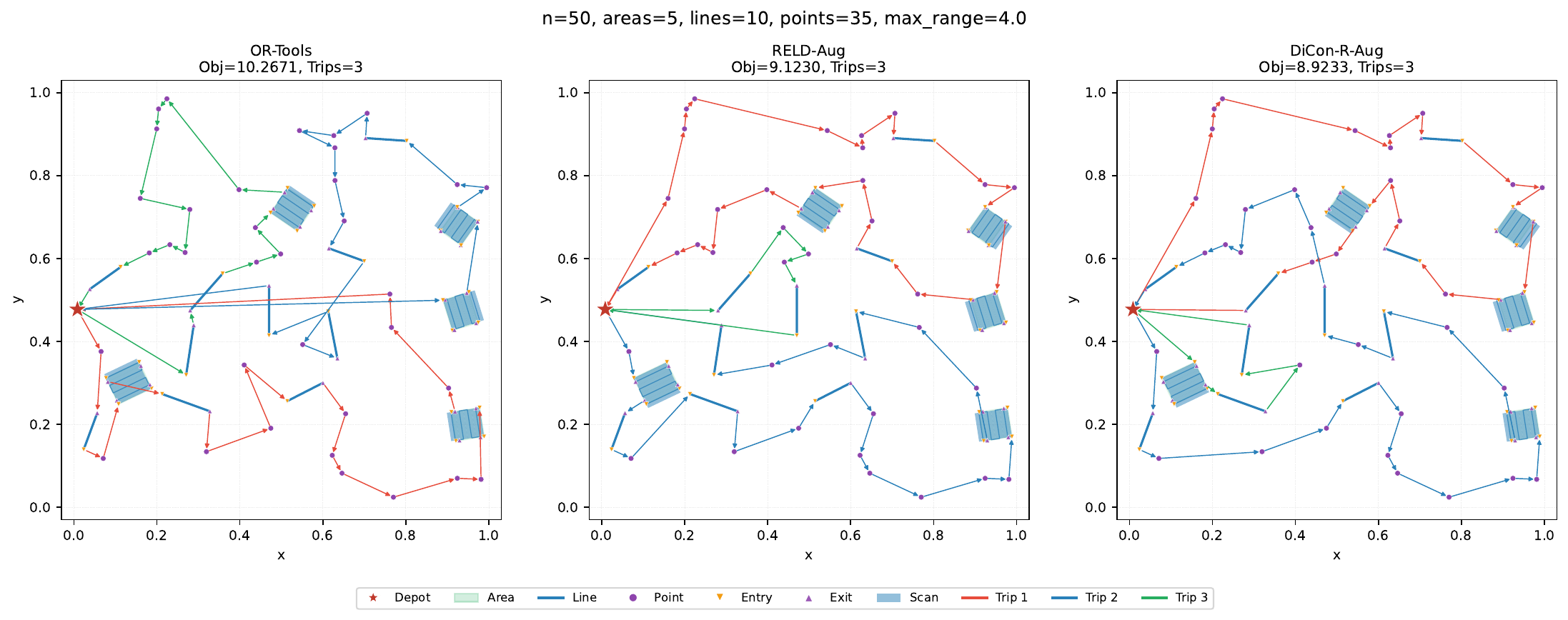}
    \caption{Performance comparison on DCCGRP}.
    \label{fig:DCCGRP}
\end{figure*}

In principle, DiCon can be extended to CGRP with more complex constraints by augmenting the encoder input with additional resource-related dimensions or introducing constraint-specific masking schemes. Since distance constraints capture a practically important deployment limitation, namely that inspection robots are often constrained by battery capacity, we focus on a distance-constrained variant of CGRP. The experimental results on distance-constrained CGRP (DCCGRP), reported in Table~\ref{tab:dcmgrp_performance} and visualized in Fig.~\ref{fig:DCCGRP}, show that our methodological contributions remain effective under this constraint setting.

\begin{table*}[htbp]
\centering
\caption{Performance comparison on DCCGRP.}
\label{tab:dcmgrp_performance}
\resizebox{0.6\linewidth}{!}{
\begin{tabular}{lcccccc}
\toprule
\multirow{2}{*}{Algorithm} & \multicolumn{2}{c}{$n=20$} & \multicolumn{2}{c}{$n=50$} & \multicolumn{2}{c}{$n=100$} \\
\cmidrule(lr){2-3} \cmidrule(lr){4-5} \cmidrule(lr){6-7}
 & Obj & Time & Obj & Time & Obj & Time \\
\midrule
OR-Tools & 6.7647 & 6.00s & 7.1064 & 1.9m  & 9.7524 & 23.2m \\
RELD    & 6.5809 & 0.09s & 6.6688 & 0.12s & 9.0431 & 0.41s \\
DiCon-R & 6.5780 & 0.09s & 6.6470 & 0.13s & 9.0197 & 0.49s \\
\bottomrule
\end{tabular}
}
\end{table*}

\section{Comparison to TSP-Trained POMO}
\label{sec:unified_encoding}

In principle, CGRP with non-point tasks can be reformulated as a node-centric problem by flattening each entry--exit pair into 2D nodes, allowing a TSP-trained solver such as POMO to be applied in a zero-shot manner. However, this reformulation requires a modified masking scheme to enforce feasibility. To examine this issue, we consider two node-centric variants. POMO-FZ applies a TSP-trained POMO in a zero-shot setting after flattening entry--exit pairs into nodes, while POMO-FR is trained on the same flattened reformulation. The results in Table~\ref{tab:pomo_dicon_comparison} show that both variants are clearly inferior to POMO trained directly on CGRP under our Unified Encoding Scheme and further behind DiCon-P. This evidence suggests that zero-shot transfer from TSP is possible only after an ad hoc reformulation, but it does not resolve the core modeling mismatch. Instead, the results support the necessity of developing neural solvers for CGRP, which is a superclass of classical routing problems such as TSP, and confirm the importance of the Unified Encoding Scheme for representing compositional geometry tasks.

\begin{table*}[htbp]
\centering
\caption{Performance comparison among zero-shot TSP-trained POMO (POMO-FZ), CGRP-trained POMO (POMO-FR), CGRP-adapted and trained POMO (POMO), and DiCon-P.}
\label{tab:pomo_dicon_comparison}
\resizebox{\linewidth}{!}{
\begin{tabular}{lccccccccccccccc}
\toprule
\multirow{2}{*}{Method} 
& \multicolumn{3}{c}{100} 
& \multicolumn{3}{c}{200} 
& \multicolumn{3}{c}{Line20} 
& \multicolumn{3}{c}{A20L30} 
& \multicolumn{3}{c}{A20L80} \\
\cmidrule(lr){2-4} 
\cmidrule(lr){5-7} 
\cmidrule(lr){8-10} 
\cmidrule(lr){11-13} 
\cmidrule(lr){14-16}
& Obj & Gap (\%) & Time 
& Obj & Gap (\%) & Time 
& Obj & Gap (\%) & Time 
& Obj & Gap (\%) & Time 
& Obj & Gap (\%) & Time \\
\midrule
POMO-FZ  & 8.544 & 2.27  & 0.26s & 14.577 & 6.58 & 1.41s & 4.426 & 1.98 & 0.04s & 13.017 & 3.94  & 0.33s & 17.387 & 5.31  & 1.16s \\
POMO-FR  & 8.557 & 2.43  & 0.26s & 14.519 & 6.16 & 1.39s & 4.355 & 0.33 & 0.05s & 13.000 & 3.80  & 0.34s & 17.459 & 5.75  & 1.17s \\
POMO     & 8.465 & 1.32  & 0.19s & 14.115 & 3.20 & 0.94s & 4.347 & 0.16 & 0.03s & 12.504 & -0.16 & 0.16s & 16.516 & 0.03  & 0.53s \\
DiCon-P  & 8.406 & 0.63  & 0.16s & 13.949 & 1.99 & 0.80s & 4.343 & 0.07 & 0.04s & 12.470 & -0.43 & 0.23s & 16.390 & -0.73 & 0.80s \\
\bottomrule
\end{tabular}
}
\end{table*}

% \begin{table*}[t]
% \centering
% \caption{\colorr{Performance comparison among zero-shot TSP-trained POMO (POMO-FZ), CGRP-trained POMO (POMO-FR), CGRP-adapted and trained POMO (POMO), and DiCon-P.}}
% \label{tab:pomo_dicon_comparison}
% \resizebox{\textwidth}{!}{
% \begin{tabular}{lcccccccccccccccccc}
% \toprule
% \multirow{2}{*}{Method} 
% & \multicolumn{3}{c}{50} 
% & \multicolumn{3}{c}{100} 
% & \multicolumn{3}{c}{200} 
% & \multicolumn{3}{c}{A0L20} 
% & \multicolumn{3}{c}{A20L30} 
% & \multicolumn{3}{c}{A20L80} \\
% \cmidrule(lr){2-4} \cmidrule(lr){5-7} \cmidrule(lr){8-10}
% \cmidrule(lr){11-13} \cmidrule(lr){14-16} \cmidrule(lr){17-19}
% & Obj & \colorr{Gap} & Time 
% & Obj & Gap & Time 
% & Obj & Gap & Time 
% & Obj & Gap & Time 
% & Obj & Gap & Time 
% & Obj & Gap & Time \\
% \midrule
% POMO-FZ 
% & 6.3074 & 1.55\% & 0.10s 
% & 8.5440 & 1.64\% & 0.26s 
% & 14.5772 & 4.70\% & 1.41s 
% & 4.4260 & 1.85\% & 0.04s 
% & 13.0169 & 4.46\% & 0.33s 
% & 17.3873 & 6.21\% & 1.16s \\

% POMO-FR 
% & 6.2636 & 0.84\% & 0.10s 
% & 8.5569 & 1.79\% & 0.26s 
% & 14.5194 & 4.29\% & 1.39s 
% & 4.3545 & 0.21\% & 0.05s 
% & 13.0004 & 4.33\% & 0.34s 
% & 17.4592 & 6.65\% & 1.17s \\

% POMO 
% & 6.2255 & 0.23\% & 0.07s 
% & 8.4646 & 0.69\% & 0.19s 
% & 14.1148 & 1.38\% & 0.94s 
% & 4.3470 & 0.04\% & 0.03s 
% & 12.5039 & 0.35\% & 0.16s 
% & 16.5157 & 0.89\% & 0.53s \\

% DiCon-P 
% & 6.2137 & 0.04\% & 0.09s 
% & 8.4064 & 0.00\% & 0.16s 
% & 13.9490 & 0.19\% & 0.80s 
% & 4.3432 & -0.05\% & 0.04s 
% & 12.4702 & 0.08\% & 0.23s 
% & 16.3903 & 0.12\% & 0.80s \\
% \bottomrule
% \end{tabular}
% }
% \end{table*}

\begin{table*}[htbp]
\centering
\caption{Comparison of CGRP instances with varying non-Point task proportions.}
\label{tab:nopoint_ratio}
\resizebox{\textwidth}{!}{
\begin{tabular}{l|ccc|ccc|ccc|ccc|ccc}
\toprule
& \multicolumn{3}{c|}{0\% (TSP)} & \multicolumn{3}{c|}{25\%} & \multicolumn{3}{c|}{50\%} & \multicolumn{3}{c|}{75\%} & \multicolumn{3}{c}{100\%} \\
    \cmidrule(lr){2-4}\cmidrule(lr){5-7}\cmidrule(lr){8-10}\cmidrule(lr){11-13}\cmidrule(lr){14-16}
    Methods& Obj & Gap(\%) & Time & Obj & Gap(\%) & Time & Obj & Gap(\%) & Time & Obj & Gap(\%) & Time & Obj & Gap(\%) & Time \\
\midrule
LKH3        & 10.629 & 0.00  & 7.43m  & 15.815 & 0.00 & 18.08m & 18.131 & 0.00  & 13.76m & 19.542 & 0.00 & 14.79m & 20.961 & 0.00 & 14.07m \\
OR-Tools    & 11.054 & 4.00  & 10.00m & 16.350 & 3.39 & 10.01m & 18.751 & 3.42  & 10.01m & 20.299 & 3.87 & 10.01m & 21.721 & 3.62 & 10.03m \\
ALNS        & 10.974 & 3.25  & 10.06m & 16.260 & 2.82 & 10.05m & 18.435 & 1.68  & 10.05m & 19.845 & 1.55 & 10.08m & 21.323 & 1.73 & 10.05m \\
\midrule
POMO        & 11.018 & 3.66  & 0.62s  & 16.297 & 3.05 & 0.28s  & 18.766 & 3.50  & 0.39s  & 20.504 & 4.92 & 0.47s  & 22.296 & 6.37 & 0.52s  \\
Dicon-P     & 10.912 & 2.66  & 0.46s  & 16.064 & 1.58 & 0.33s  & 18.387 & 1.41  & 0.41s  & 19.979 & 2.23 & 0.54s  & 21.578 & 2.94 & 0.61s  \\
RELD        & 10.849 & 2.07  & 0.54s  & 16.022 & 1.31 & 0.28s  & 18.327 & 1.08  & 0.31s  & 19.864 & 1.64 & 0.39s  & 21.384 & 2.02 & 0.46s  \\
Dicon-R     & 10.819 & 1.79  & 0.59s  & 15.904 & 0.56 & 0.36s  & 18.195 & 0.35  & 0.44s  & 19.661 & 0.61 & 0.57s  & 21.180 & 1.04 & 0.66s  \\
\midrule
POMO-Aug    & 10.910 & 2.64  & 4.97s  & 16.088 & 1.73 & 1.86s  & 18.514 & 2.12  & 3.01s  & 20.266 & 3.70 & 3.50s  & 22.032 & 5.11 & 3.63s  \\
Dicon-P-Aug & 10.823 & 1.83  & 3.99s  & 15.947 & 0.84 & 2.56s  & 18.241 & 0.61  & 3.83s  & 19.792 & 1.28 & 4.72s  & 21.384 & 2.02 & 5.33s  \\
RELD-Aug    & 10.784 & 1.46  & 3.00s  & 15.906 & 0.58 & 2.20s  & 18.159 & 0.16  & 2.41s  & 19.700 & 0.81 & 3.87s  & 21.198 & 1.13 & 4.24s  \\
Dicon-R-Aug & 10.750 & 1.14  & 4.03s  & 15.839 & 0.15 & 2.90s  & 18.057 & -0.40 & 4.12s  & 19.558 & 0.08 & 5.04s  & 21.007 & 0.22 & 5.67s  \\
\bottomrule
\end{tabular}
}
\end{table*}

\section{Impact of Non-Point Task Proportion on CGRP Solvers}
% \colorb{Add analysis on the task proportion effect on traditional and neural solvers.}

To examine solver robustness under varying task geometries, we generate CGRP instances with 200 tasks and vary the proportion of non-point tasks (i.e., lines or areas) from 0\% (pure TSP) to 100\%, with increments of 25\%. For each setting, we sample 20 instances and report results in Table~\ref{tab:nopoint_ratio}. DiCon-R and DiCon-P consistently outperform the neural baselines RELD and POMO across all settings, and the performance gap widens as the proportion of non-point tasks increases—highlighting the geometric adaptability of our model. OR-Tools remains stable, while ALNS improves with higher non-point proportions and surpasses RELD when the ratio exceeds 75\%, demonstrating its flexibility in handling non-point tasks. Overall, DiCon shows strong compositional generalization and is especially effective in non-standard routing scenarios.

\section{Entropy of MHA and MHDA} \label{sec: entropy}

\begin{figure*}
    % \vspace{1mm}
    \centering
    \includegraphics[width=0.6\columnwidth]{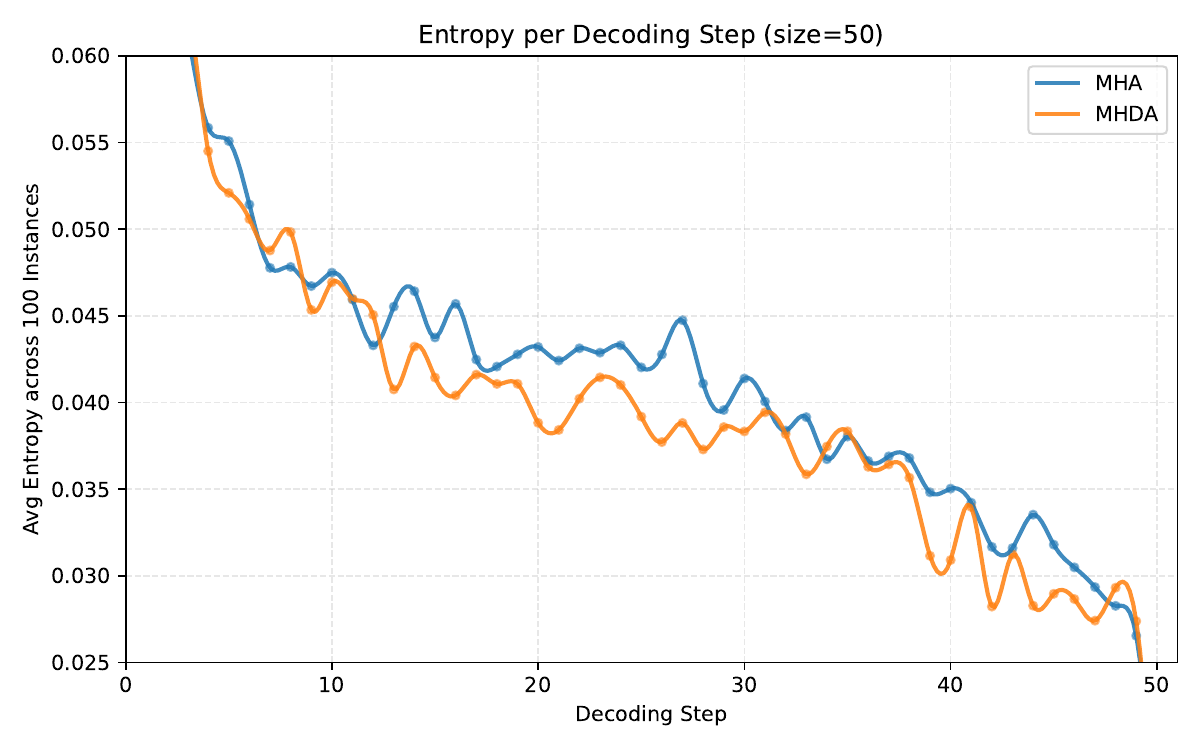}
    % \vspace{-5mm}
    \caption{Entropy comparison between MHA and MHDA.
    }
    \label{fig:entropy}
    \vspace{-2mm}
\end{figure*}

To further demonstrate the effectiveness of differential attention (DA), we compare standard MHA with the proposed multi-head differential attention (MHDA) integrated into the RELD backbone. Specifically, we compute the output entropy at each decoding step on 100 randomly sampled CGRP50 instances for both RELD (with MHA) and RELD with MHDA. As illustrated in Fig.~\ref{fig:entropy}, MHDA consistently yields lower entropy across the decision steps compared to MHA, indicating more confident and focused policy outputs. The reduction is generally consistent, with an absolute entropy drop of about $0.001$--$0.004$ for most steps and occasionally close to $0.005$ in later decoding stages, relative to a baseline entropy level of approximately $0.03$--$0.06$. This observation aligns with the intended effect of DA, which adaptively amplifies semantically critical ones, thereby enhancing decision consistency across different geometry-task compositions.

\begin{table*}[htbp]
\centering
\caption{The decoder parameter amount in RELD and DiCon-R.}
\label{tab:decoder_params}
\begin{tabular}{lccc}
\toprule
Component & RELD & DiCon-R & Difference \\
\midrule
Decoder Total Params & 213,760 & 262,920 & +49,160 \\
\bottomrule
\end{tabular}
\end{table*}

\begin{table*}[t]
\centering
\caption{Results of POMO, RELD, and their DA versions.}
\label{tab:da_results}
\resizebox{\linewidth}{!}{
\begin{tabular}{lccccccccccccccc}
\toprule
& \multicolumn{3}{c}{100} 
& \multicolumn{3}{c}{200} 
& \multicolumn{3}{c}{Line20} 
& \multicolumn{3}{c}{A20L30} 
& \multicolumn{3}{c}{A20L80} \\
\cmidrule(lr){2-4} 
\cmidrule(lr){5-7} 
\cmidrule(lr){8-10} 
\cmidrule(lr){11-13} 
\cmidrule(lr){14-16}
& Obj & Gap (\%) & Time 
& Obj & Gap (\%) & Time 
& Obj & Gap (\%) & Time 
& Obj & Gap (\%) & Time 
& Obj & Gap (\%) & Time \\
\midrule
POMO        & 8.465 & 1.32 & 0.19s & 14.115 & 3.20 & 0.94s & 4.347 & 0.16 & 0.03s & 12.504 & -0.16 & 0.16s & 16.516 & 0.03  & 0.53s \\
POMO\_DA    & 8.431 & 0.92 & 0.16s & 14.024 & 2.54 & 0.79s & 4.345 & 0.12 & 0.03s & 12.493 & -0.25 & 0.23s & 16.463 & -0.28 & 0.43s \\
RELD        & 8.406 & 0.63 & 0.26s & 13.922 & 1.79 & 1.17s & 4.345 & 0.12 & 0.03s & 12.461 & -0.50 & 0.19s & 16.370 & -0.85 & 1.01s \\
RELD\_DA    & 8.403 & 0.59 & 0.36s & 13.892 & 1.57 & 1.74s & 4.342 & 0.05 & 0.05s & 12.456 & -0.54 & 0.27s & 16.360 & -0.91 & 1.32s \\
RELD\_double & 8.405 & 0.62 & 0.27s & 13.898 & 1.61 & 1.25s & 4.344 & 0.08 & 0.04s & 12.458 & -0.53 & 0.21s & 16.371 & -0.84 & 0.68s \\
\bottomrule
\end{tabular}
}
\end{table*}

\section{Efficiency Analysis of DA}

We analyzed the parameter overhead of DA and reported the statistics in Table~~\ref{tab:decoder_params}. DA adds about $23\%$ more parameters to the decoder relative to RELD. However, since RELD itself uses a lightweight decoder, this corresponds to only $3.5\%$ additional parameters for the entire model. Moreover, Table~\ref{tab:da_results} shows that DA introduces only a slight runtime increase.

More importantly, the key issue is not only how many parameters are added, but also where and how they are introduced. To examine whether the gain simply comes from a larger decoder, we introduce a control baseline, RELD\_double, whose decoder parameter budget is matched to RELD\_DA. Specifically, RELD\_double adds an adapter consisting of a three-layer MLP with a hidden dimension of $128$, together with a head-wise gate after the feed-forward layer and a residual connection. As reported in Table~\ref{tab:da_results}, RELD\_double consistently underperforms RELD\_DA and is even worse than RELD on CGRP50.

\section{Experimental Details}
\subsection{Hyperparameter of Conventional Baselines}
\textbf{LKH3.} 
The LKH-3 heuristic leverages the open-source LKH-3 solver, a state-of-the-art implementation for solving large-scale TSP variants using Lin-Kernighan style $k$-opt moves with candidate edge pruning and sequential search. The algorithm employs a multiple restart strategy (we set it to 20 in the main experiments) that initiates independent alternating optimization cycles from different starting configurations. Each restart consists of three stages: node selection, route optimization, and node choice refinement. The first restart uses a deterministic greedy nearest-neighbor strategy for node selection to ensure a high-quality baseline, while subsequent restarts introduce increasing randomization to explore diverse regions of the solution space. Given the selected nodes, the algorithm constructs an ATSP subproblem with explicit distance matrices (exit-to-entry Euclidean distances) and invokes LKH-3 to solve the routing order, exploiting its powerful 5-opt sequential moves and sensitivity-based candidate edge filtering. Following route optimization, a node choice refinement pass fixes the visiting sequence and re-evaluates all candidate nodes for each task to minimize local travel cost, iterating for up to 5 passes. The alternating optimization between LKH-3 routing and node choice refinement repeats for 3 iterations within each restart, allowing the two decision components to mutually improve. The algorithm retains the solution with the minimum objective value across all independent restarts. We further evaluate the effect of the number of restarts. The results are reported in Table~\ref{tab:lkh_dicon_results}. As the number of restarts increases, the solution quality improves, but the runtime also increases substantially.

\begin{table*}[t]
\centering
\caption{Results of LKH3-20, LKH3-100, LKH3-1000, DiCon-R, and DiCon-R-Aug. Here, 20 denotes the number of restarts.}
\label{tab:lkh_dicon_results}
\resizebox{\linewidth}{!}{
\begin{tabular}{lccccccccccccccc}
\toprule
\multirow{2}{*}{Methods}
& \multicolumn{3}{c}{100}
& \multicolumn{3}{c}{200}
& \multicolumn{3}{c}{Line20}
& \multicolumn{3}{c}{A20L30}
& \multicolumn{3}{c}{A20L80} \\
\cmidrule(lr){2-4}
\cmidrule(lr){5-7}
\cmidrule(lr){8-10}
\cmidrule(lr){11-13}
\cmidrule(lr){14-16}
& Obj & Gap (\%) & Time
& Obj & Gap (\%) & Time
& Obj & Gap (\%) & Time
& Obj & Gap (\%) & Time
& Obj & Gap (\%) & Time \\
\midrule
LKH3-20      & 8.354 & 0.00  & 50.30m & 13.677 & 0.00  & 3.32h  & 4.340 & 0.00  & 8.57m  & 12.524 & 0.00  & 16.00m & 16.510 & 0.00  & 43.63m \\
LKH3-100     & 8.346 & -0.10 & 3.51h  & 13.631 & -0.34 & 13.45h & 4.339 & -0.02 & 18.17m & 12.449 & -0.60 & 40.53m & 16.401 & -0.66 & 2.57h  \\
LKH3-1000    & 8.345 & -0.11 & 6.67h  & 13.627 & -0.37 & 16.69h & 4.339 & -0.02 & 1.66h  & 12.385 & -1.11 & 6.62h  & 16.318 & -1.16 & 16.67h \\
DiCon-R      & 8.388 & 0.41  & 0.31s  & 13.845 & 1.23  & 1.83s  & 4.341 & 0.02  & 0.05s  & 12.437 & -0.70 & 0.27s  & 16.303 & -1.25 & 1.37s  \\
DiCon-R-Aug  & 8.363 & 0.10  & 4.01s  & 13.755 & 0.57  & 20.95s & 4.339 & -0.02 & 0.31s  & 12.385 & -1.11 & 3.66s  & 16.217 & -1.78 & 10.87s \\
\bottomrule
\end{tabular}
}
\end{table*}

\textbf{OR-Tools.} 
We employ Google's OR-Tools Routing Library (v9.15) to solve the HTTSP. The problem is modeled as a single-vehicle routing problem over all $M+1$ candidate nodes (including the depot). To handle the heterogeneous node selection, we formulate \emph{disjunction constraints}: for each task $i$ with candidate node set $\mathcal{V}_i$, we add a disjunction group with maximum cardinality 1, requiring exactly one node from $\mathcal{V}_i$ to be visited. A sufficiently large penalty ($2M \times 10^5$) is assigned to each disjunction to ensure full task coverage. The asymmetric travel cost matrix is defined by the Euclidean distance from each node's exit point to the next node's entry point, scaled to integer precision with a factor of $10^5$ as required by the solver's integer arithmetic. The initial solution is constructed using the \texttt{PATH\_CHEAPEST\_ARC} strategy, which greedily extends the route by selecting the minimum-cost arc from the current node. The solution is then improved by the \texttt{Guided\_Local\_Search (GLS)} metaheuristic, which augments the objective with penalties on arcs that frequently appear in local optima, enabling the solver to escape plateaus. The local search phase internally combines multiple neighborhood operators including 2-opt, Or-opt, Relocate, and Cross-exchange moves, managed by the solver's C++ engine.

\textbf{ALNS Algorithm.}
The ALNS operates for a maximum of 500 iterations to balance exploration and exploitation within the given time budget. The simulated annealing acceptance criterion begins with an initial temperature set to $\mathbb{T}_0 = 0.05 \times f(\hat{s}_0)$, where $f(\hat{s}_0)$ denotes the objective value of the initial solution, providing moderate diversification at the start, and employs a geometric cooling schedule with rate $\tau = 0.98$ such that $\mathbb{T}_{k+1} = \tau \cdot \mathbb{T}_k$ to gradually intensify the search toward high-quality regions. The destroy ratio is configured at $\rho = 0.3$, removing approximately 30\% of nodes in each iteration to maintain a balance between preserving solution structure and enabling substantial modifications. The algorithm employs three destroy operators: random removal for unbiased exploration, worst removal targeting high-cost nodes, and related removal, which selects a seed node and removes its spatially proximate neighbors to enable regional route reconfiguration. Two repair operators are implemented: greedy insertion for quick feasible reconstruction and regret-based insertion for quality-conscious restoration. Operator selection follows a roulette wheel mechanism based on adaptive weights that are recalibrated every 100 iterations using a reaction factor of $\phi = 0.1$ according to $w_i^{new} = (1-\phi) w_i^{old} + \phi \bar{s}_i$, where $\bar{s}_i$ represents the average score of operator $i$, gradually favoring operators that consistently discover improved solutions. To further exploit the flexibility of heterogeneous task structures, the algorithm invokes node choice optimization every 10 iterations to refine entry-exit point selections for line and area tasks while maintaining computational efficiency throughout the search process.

\begin{table*}[htbp]
\centering
\caption{Results with different augmentation settings.}
\label{tab: aug_ablation}
\begin{tabular}{l|ccccccc}
\toprule
 & 20 & 50 & 100 & 150 & 200 & 300 \\
\midrule
$\mathbb{A}$=2 & 6.0726 & \textbf{6.2073} & 8.3921 & 10.6693 & 13.8602 & 15.7547 \\
$\mathbb{A}$=3 & 6.0728 & 6.2079 & 8.3923 & 10.6633 & 13.8512 & 15.7267 \\
$\mathbb{A}$=4 (Ours) & \textbf{6.0722} & 6.2090 & \textbf{8.3879} & \textbf{10.6594} & \textbf{13.8449} & \textbf{15.7022} \\
\bottomrule
\end{tabular}
\end{table*}

\begin{table*}[htbp]
\centering
\caption{Results with different $\lambda_{ins}$ settings.}
\label{tab: lambda_ablation}
\begin{tabular}{l|ccccccc}
\toprule
 & 20 & 50 & 100 & 150 & 200 & 300 \\
\midrule
$\lambda_{ins}$=0.05 & 6.0744 & \textbf{6.2088} & 8.3889 & 10.6581 & 13.8396 & 15.7209 \\
$\lambda_{ins}$=0.10 (Ours) & \textbf{6.0722} & 6.2090 & \textbf{8.3879} & \textbf{10.6594} & \textbf{13.8449} & \textbf{15.7022} \\
$\lambda_{ins}$=0.20 & 6.0733 & 6.2091 & 8.3898 & 10.6552 & 13.8380 & 15.7391 \\
\bottomrule
\end{tabular}
\vspace{-2mm}
\end{table*}

\subsection{Hyperparameter Study}
\label{sec:CL_hyperparam}
We evaluate the sensitivity of DiCon to key hyperparameters, including the number of augmentations $\mathbb{A}$, the instance-level contrastive loss weight $\lambda_{ins}$, and the intra-task contrastive loss weight $\lambda_{intra}$.

\noindent\textbf{Effect of Augmentation Number $\mathbb{A}$.}
Table~\ref{tab: aug_ablation} shows that increasing $\mathbb{A}$ improves performance consistently. Using $\mathbb{A}$ achieves the best overall results across all instance sizes, supporting our claim that strong data augmentations help the model learn geometry-invariant representations.

\noindent\textbf{Effect of Instance-Level Contrastive Loss Weight $\lambda_{\text{ins}}$.}
As shown in Table~\ref{tab: lambda_ablation}, setting $\lambda_{ins}{=}0.10$ yields the best performance. A smaller value (e.g., $0.05$) leads to insufficient contrastive pressure, while a larger one (e.g., $0.20$) may distort the optimization landscape.

\noindent\textbf{Effect of Intra-Task Contrastive Loss Weight $\lambda_{\text{intra}}$.}
Table~\ref{tab: w_struct_ablation} indicates that $\lambda_{intra}{=}0.02$ performs best overall. Similar to $\lambda_{ins}$, both smaller and larger values degrade performance, suggesting that balancing intra-task discrimination is key to generalizable representations.

\begin{table*}[htbp]
\vspace{-2mm}
\centering
\caption{Results with different $\lambda_{intra}$ settings.}
\label{tab: w_struct_ablation}
\begin{tabular}{l|ccccccc}
\toprule
 & 20 & 50 & 100 & 150 & 200 & 300 \\
\midrule
$\lambda_{intra}$=0.01 & 6.0724 & 6.2080 & 8.3905 & \textbf{10.6498} & \textbf{13.8433} & 15.7511 \\
$\lambda_{intra}$=0.02 (Ours) & 
\textbf{6.0722} & 6.2090 & \textbf{8.3879} & 10.6594 & 13.8449 & \textbf{15.7022} \\
$\lambda_{intra}$=0.03 & 6.0734 & \textbf{6.2066} & 8.3886 & 10.6578 & 13.8473 & 15.7137 \\
\bottomrule
\end{tabular}
\vspace{-2mm}
\end{table*}

% \begin{table*}[htbp]
%   \centering
%   % \setlength\tabcolsep{9pt}
%   \caption{Used assets, licenses, and their usage.}
%   \vspace{4pt}
%   \begin{threeparttable}
%     \resizebox{0.9\textwidth}{!}{
%     \begin{tabular}{c|c|c|c}
%     \toprule
%     \toprule
%     % \midrule
%     \textbf{Type}  & \textbf{Asset} & \textbf{License} & \textbf{Usage} \\
%     \midrule
%     % \midrule
%     \multirow{4}[2]{*}{Code} & LKH3 \cite{helsgaun2017lkh3} & Available for academic use & Evaluation \\
%     & OR-Tools~\cite{ortools_routing} & Apache-2.0 license & Evaluation \\
%      & POMO \cite{kwon2020pomo} &  MIT License & Revision \\
%       & RELD \cite{huang2025rethinking} &  No license (assumed all rights reserved) & Revision \\
%     \bottomrule
%     \end{tabular}}
%     \end{threeparttable}
%   \label{tab:asset}%
% \end{table*}%

\section{Broader Impacts}
\label{appendix:impact}
DiCon may provide positive societal value by improving the efficiency and scalability of routing systems under the unified CGRP formulation, which covers point-only, line-only, area-only, and hybrid task geometries. These settings arise in logistics, robotic inspection, disaster response, infrastructure monitoring, and agricultural surveying. By handling these routing scenarios within a unified neural solver, DiCon can reduce problem-specific redesign and support faster planning in large or compositionally novel environments. Its differential attention and double-level contrastive learning components further improve decision-making and representation learning, helping produce high-quality routes with lower computational cost than exact or traditional heuristic solvers.

DiCon should be used as a decision-support tool rather than a replacement for human oversight. In safety-critical applications, generated routes should be validated against operational constraints, environmental uncertainty, and domain-specific safety requirements before deployment.

% \section{Licenses for Existing Assets}
% \label{appendix:assets}

% The used assets in this work are listed in Table~\ref{tab:asset}. Where applicable, we reference publicly available implementations for evaluation or reproduction purposes.

%%%%%%%%%%%%%%%%%%%%%%%%%%%%%%%%%%%%%%%%%%%%%%%%%%%%%%%%%%%%

% \newpage
% \input{checklist.tex}

\end{document}